\documentclass[graybox]{svmult}

\usepackage[numbers,sort&compress]{natbib}
\usepackage{graphicx}
\usepackage{float,array}
\usepackage{amsmath}

\usepackage{amsthm}
\usepackage{float}
\usepackage{amssymb}
\usepackage{algpseudocode}
\usepackage{algorithm}
\usepackage{epstopdf}
\usepackage{longtable}

\usepackage{multirow}
\usepackage{caption}
\usepackage{subfig}
\usepackage{balance}
\usepackage{color}
\usepackage[normalem]{ulem}
\newcommand{\todo}[1]{{\color{black} #1}}
\newcommand{\rev}[1]{{\color{black} #1}}

\newcommand{\et}{\emph{et al.} }

\usepackage{ifsym}

\makeindex             % used for the subject index
\begin{document}

\title*{Robotic Wireless Sensor Networks}
\author{Pradipta Ghosh*, Andrea Gasparri, Jiong Jin and Bhaskar Krishnamachari}
\institute{Pradipta Ghosh (*Corresponding author) \at University of Southern California, Los Angeles, CA-90089 \email{pradiptg@usc.edu}
\and Andrea Gasparri \at Università degli studi "Roma Tre", Via della Vasca Navale, 79
00146 Roma (Italy) \email{gasparri@dia.uniroma3.it}
\and Jiong Jin \at Swinburne University of Technology, Melbourne, VIC 3122, Australia \email{jiongjin@swin.edu.au}
\and Bhaskar Krishnamachari \at University of Southern California, Los Angeles, CA-90089 \email{bkrishna@usc.edu}}
\maketitle

% \abstract*{Each chapter should be preceded by an abstract (10--15 lines long) that summarizes the content. The abstract will appear \textit{online} at \url{www.SpringerLink.com} and be available with unrestricted access. This allows unregistered users to read the abstract as a teaser for the complete chapter. As a general rule the abstracts will not appear in the printed version of your book unless it is the style of your particular book or that of the series to which your book belongs.
% Please use the 'starred' version of the new Springer \texttt{abstract} command for typesetting the text of the online abstracts (cf. source file of this chapter template \texttt{abstract}) and include them with the source files of your manuscript. Use the plain \texttt{abstract} command if the abstract is also to appear in the printed version of the book.}

\abstract{
In this chapter, we present a literature survey of an emerging, cutting-edge, and multi-disciplinary field of research at the intersection of Robotics and Wireless Sensor Networks (WSN) which we refer to as \emph{Robotic Wireless Sensor Networks (RWSN)}. 
%We define a \emph{Robotic Wireless Sensor Network} as an autonomous networked multi-robot system that is able to cooperatively learn and adapt towards achieving certain \emph{sensing goals} and is able to maintain certain \emph{communication performance requirements}. 
We define a \emph{RWSN} as an autonomous networked multi-robot system that aims to achieve certain \emph{sensing goals} while meeting and maintaining certain \emph{communication performance requirements}, through cooperative control, learning and adaptation. 
While both of the component areas, i.e., Robotics and WSN, are very well-known and well-explored, there exist a whole set of new opportunities and  research directions at the intersection of these two fields which are relatively or even completely unexplored. One such example would be the use of a set of robotic routers to set up a temporary communication path between a sender and a receiver that uses the controlled mobility to the advantage of packet routing. We find that there exist only a limited number of articles to be directly categorized as RWSN related works whereas there exist a range of articles in the robotics and the WSN literature that are also relevant to this new field of research. 
To connect the dots, we first identify the core problems and research trends related to RWSN such as connectivity, localization, routing, and robust flow of information. Next, we classify the existing research on RWSN as well as the relevant state-of-the-arts from robotics and WSN community according to the problems and trends identified in the first step.
Lastly, we analyze what is missing in the existing literature, and identify topics that require more research attention in the future.
}

\section{Introduction}
\label{sec:intro}
Robotics has been a very important and active field of research over last couple of decades with the main focus on seamless integration of robots in human lives to assist and to help human in difficult, cumbersome jobs such as search and rescue in disastrous environments and exploration of unknown environments~\cite{penders2011robot,murphy2004trial}. 
The rapid technological advancements over last two decades in terms of cheap and scalable hardware with necessary software stacks, have provided a huge momentum to this field of research.
% In spite of a huge amount of research in this field over last few decades,  Robotics still  lacks sufficient integration in human life. 
As part of this increasing stream of investigations into robotics, researchers have been motivated to look into the collaborative aspects where a group of robots can work in synergy to perform a set of diverse tasks~\cite{gazi2011swarm,Winfield:2008}. 
Nonetheless, most of the research works on collaborative robotics, such as swarming, have remained mostly either theoretical concepts or incomplete practical systems which lack some very important pieces of the puzzle such as realistic communication channel modeling and efficient network protocols for interaction among the robots. \rev{Note that, we use the term ``realistic communication channel model'' to refer to a wireless channel model that accounts for most of the well-known dynamics of a standard wireless channel such as path loss, fading, and shadowing~\cite{rappaport1996wireless}.}
On the other hand, the field of Wireless Networks (more specifically, Wireless Sensor Networks (WSN) and Wireless AdHoc Networks) has been explored extensively by communication and network researchers where the nodes are considered static (Sensor nodes) or mobile without control (Mobile AdHoc Network). With the availability of cheap easily programmable robots, researchers have started to explore the advantages and opportunities granted by the controlled mobility in the context of Wireless Networks. Nonetheless, the mobility models used by the network researchers remained simple and impractical, and not very pertinent to robotic motion control until last decade. 
% Over last few decade researcher have achieved exceptional advancements in the area of communication, mainly Wireless Network, in aspect such as portability, reliability, sophistication and cost effectiveness. 
%In current scenario, a day without internet, cell phones etc are unimaginable. 
%On the other hand Robots are not that much integrated in our life today just like cell-phones, internet wasn't few decades ago. But that also depend on the definition of  robot. While the Encyclopaedia Britannica  defines robot as \textit{``any automatically operated machine that replaces human effort, though it may not resemble human beings in appearance or perform functions in a humanlike manner.''} some scientists also describe a robot as \textit{``a machine that automatically performs complicated often repetitive tasks.''} In the light of second definition many instrument used by us already falls under the domain of Robotics.  However, Robots as per all definition is envisioned as a integrate part of our life in recent future by many groups of people including researchers, writers etc. 
%Robots have been already applied successfully in a large number of domains such as rescue operations, fire fighting, underground mining, exploration, robot sports, etc.\\
% Over last decade, most of the researchers focused on those fields individually.

Over last decade, a handful of researchers noticed the significant disconnection between the robotics and the wireless network research communities and its bottleneck effects in the full fledged development of a network of collaborative robots. 
Consequently, researchers have tried to incorporate wireless network technologies in robotics and vice verse, which opened up a whole new field of research at the intersection of robotics and wireless networks.
% As the main step towards the integration of both research fields, many researchers equipped robots with wireless communication capabilities, stimulated by the observation that multi-robot systems tend to have several advantages over their single-robot counterparts.
% The choice of Wireless technologies over wired technology is quite obvious. 
% The main linking element of this process is Wireless Sensor Network (WSN) that can be defined as a group of small devices with sensing, wireless communication and computation ability, intended to monitor or gather information in many diverse situations. 
This new research domain is called by many different names such as \emph{``Wireless Robotics Networks''}, \emph{ ``Wireless Automated Networks''} and \emph{``Networked Robots''}. In this chapter, keeping in mind that the primary task of teams of robots in many application context might be pure sensing, we will refer to this field as \emph{``Robotic Wireless Sensor Networks (RWSN)''}.  
%Before going any further let us present some definitions related of this new field, provided by different group of researchers.
According to the IEEE Society of Robotics and Automation's Technical Committee: \emph{``A `networked robot' is a robotic device connected to a communications network such as the Internet or LAN. The network could be wired or wireless, and based on any of a variety of protocols such as TCP, UDP, or 802.11. Many new applications are now being developed ranging from automation to exploration. There are two subclasses of Networked Robots: (1) Tele-operated, where human supervisors send commands and receive feedback via the network. Such systems support research, education, and public awareness by making valuable resources accessible to broad audiences; (2) Autonomous, where robots and sensors exchange data via the network. In such systems, the sensor network extends the effective sensing range of the robots, allowing them to communicate with each other over long distances to coordinate their activity. The robots in turn can deploy, repair, and maintain the sensor network to increase its longevity, and utility. A broad challenge is to develop a science base that couples communication to control to enable such new capabilities.'' } 
We define a \emph{RWSN} as an autonomous networked multi-robot system that aims to achieve certain \emph{sensing goals} while meeting and maintaining certain \emph{communication performance requirements} via cooperative control, learning, and adaptation. 
%We define a \emph{Robotic Wireless Sensor Network} as an autonomous networked multi-robot system that is able to cooperatively learn and adapt towards achieving certain \emph{sensing goals} and is able to maintain certain \emph{communication performance requirements}. 
Another important definition related to this field is \emph{``cooperative behavior''} which is defined as follows: \emph{``given some task specified by a designer, a multiple-robot system displays cooperative behavior if, due to some underlying mechanism (i.e., the ``mechanism of cooperation''), there is an increase in the total utility of the system.''} A group of cooperative robots is of more interest than single robot because of some fundamental practical reasons such as easier completion and performance benefits of using multiple simple, cheap, and flexible robots for complex tasks. 
% For the rest of the manuscript we will mention the keyword \emph{``Wireless Networked Robots (RWSN)\,''} to indicate our field of interest. 
 
Over the years, robotics and wireless network researchers have developed algorithms as well as hardware solutions that directly or indirectly fall under the umbrella of RWSN.
Network of robots are already experimentally applied and tested in a range of applications such as Urban Search And Rescue missions (USAR), fire fighting, underground mining, and exploration of unknown environments.
The very first practical USAR missions was launched during the rescue operations at the World Trade Center on 11 September 2001~\cite{murphy2004trial} using a team of four robots. 
In the context of fire fighting, a group of Unmanned Aerial Vehicles (UAV) was used for assistance at the Gestosa (Portugal) forest fire in May 2003 by Ollero \et \cite{ollero2003helicopter}. 
Communication links between the robots in such contexts can be very dynamic and unreliable, thereby, require special attention for an efficient operation. 
This requires careful movement control by maintaining good link qualities among the robots.
Among other application contexts, underground mining is very improtant. Due to many difficulties like lack of accurate maps, lack of structural soundness, harshness of the environment (e.g. low oxygen level), and the danger of explosion of methane, accidents are almost inevitable in underground mining resulting in the deaths of many mine workers.
Thrun \et \cite{thrun2004autonomous} developed a robotic system that can autonomously explore and acquire three-dimensional maps of abandoned mines.
Later, Murphy \et \cite{murphy2009mobile} and Weiss \et \cite{weiss2008statistical}, also presented models and techniques for using a group of robots in underground mining.
% Such system can be used to guide and protect the miners
The field of cooperative autonomous driving, one of the major research focus in the automobile industry, also falls under the broad umbrella of RWSN. 
Baber \et \cite{baber2005cooperative}, Nagel \et \cite{nagel2007intelligent}, Milanes \et \cite{milanes2011cooperative}, and  Xiong \et \cite{xiong2010resilient} worked on solving a range of problems in practical implementations of autonomous driving systems and flocking of multiple unmanned vehicles like Personal Air Vehicle (PAV).
Robot swarms~\cite{parker1998alliance, ibach2005cero}, which deal with large numbers of autonomous and homogeneous robots, are also special cases of RWSN. Swarms have limited memory and have very limited self-control capabilities.
Example use cases of swarms are in searching and collecting tasks (food harvesting \cite{kovacs2009connectivity}, in collecting rock samples on distant planets \cite{steels1990cooperation}), or in collective transport of palletized loads \cite{stilwell1993toward}.
Penders \et \cite{penders2011robot}, developed a robot swarm to support human in search missions by surrounding them and continuously sensing and scanning the surroundings to inform them about potential hazards. 
A swarm of robots can also be used in future health-care systems e.g., swarm of micro robots can be used to identify and destroy tumor/cancer cells.
Military application is another obvious field of application. 
Many researchers have been working on developing military teams of autonomous UAVs, tanks and Robots, e.g., use of Unmanned Ground Vehicles (UGV) during RSTA missions. 
One example of such project is ``Mobile Autonomous Robot Systems (MARS),'' sponsored by DARPA. 
The works of Nguyen \et \cite{nguyen2003autonomous} and Hsieh \et \cite{hsieh2007adaptive} are mentionable on military application of Networked Robots.
In the field of Exploration, the most famous example is the twin Mars Exploration Rover (MER) vehicles. They landed on Mars in the course of January 2004 \cite{erickson2006living}. 
Among other applications, hazardous waste management, robot sports \cite{calkins2011overview}, mobile health-care \cite{petelin2007deployment}, smart home \cite{baeg2007robomaidhome, baeg2007building,de2011robots}, smart antenna, deployment of communication network and improvement of current communication infrastructure \cite{correll2009ad,batalin2004coverage}, and cloud networked robotics \cite{du2011design,quintas2011cloud,kamei2012cloud} are also important. In summary, there exists a huge range of applications of a RWSN. In Table~\ref{app_summ}, we list the different types of applications of a RWSN.
%\todo{talk briefly about two different class of swarm robotics}.

\begin{table}
\centering
\caption{Application Summary of RWSN}
\label{app_summ}
\begin{tabular}{|c|c|}
\hline
Applications & Reference \\ \hline \hline
Search and Rescue & \cite{baber2005cooperative,murphy2004trial,ollero2003helicopter,penders2011robot}\\ \hline
Mining & \cite{thrun2004autonomous,murphy2009mobile,weiss2008statistical} \\ \hline
Autonomous Driving & \cite{baber2005cooperative,nagel2007intelligent,milanes2011cooperative,xiong2010resilient} \\ \hline
Robot Swarm & \cite{parker1998alliance,ibach2005cero,kovacs2009connectivity,steels1990cooperation,stilwell1993toward} \\ \hline
Military Applications & \cite{nguyen2003autonomous,hsieh2007adaptive,erickson2006living} \\ \hline
Robot sports & \cite{calkins2011overview} \\ \hline
Mobile health-care & \cite{petelin2007deployment} \\ \hline
Smart Home  & \cite{baeg2007robomaidhome, baeg2007building,de2011robots} \\ \hline
Deployment of Communication Network & \cite{correll2009ad,batalin2004coverage} \\ \hline
Cloud Networked Robotics & \cite{du2011design,quintas2011cloud,kamei2012cloud} \\ \hline
\end{tabular}
\end{table}

In this chapter, we present a literature survey of the existing state-of-the-arts on RWSN. 
We discover that while there exist significant amount of works~\cite{DistCtrlRobotNetw} in both the ancestral fields (Robotics and WSN) that are also relevant to RWSN, most of these state-of-the-arts cannot be directly classified as RWSN related works, yet, should not be omitted. 
Nonetheless, there also exist a range of works that properly lie at the intersection of robotics and WSN and, therefore, directly fall under the purview of RWSN. 
To draw a complete picture of the RWSN related state-of-the-arts, we \textbf{first} identify and point out current research problems, trends, and challenges in the field of RWSN such as connectivity, localization, routing, and robust flow of information. 
While some of these problems are inherited from the fields of robotics (such as path control and coordination) and WSN (such as routing and localization), a new class of independent problems have also emerged such as link quality maintenance, radio signal strength information (RSSI) estimations in present and future location of a robot, and guaranteed proximity between neighboring nodes. 
\textbf{Secondly,} we categorize the existing research in accordance to the problems they address.
In doing so, we also include solutions that are not directly applicable but provide with a solution base to build upon. 
For example, in the contexts of connectivity maintenance, the traditional solutions involve representing the network as a graph by employing simple unit disk wireless connectivity model~\cite{yang2010decentralized}.
However, in practical employments, the wireless communication links do not follow unit-disk model and rather follow a randomized fading and shadowing model~\cite{rappaport1996wireless}. 
Thus, such existing solutions do not directly fall under RWSN yet can be modified to include more realistic communication model and, thus, should not be omitted. 
\textbf{Lastly,} we discuss what is missing in the existing literature, if any, as well as some potential directions of future research in the field of RWSN.  
\emph{In this chapter, we further discuss how and where do the existing works fit in the context of the layered architecture (Internet model) of a network stack in order to identify key networking goals and problems in a RWSN.}
%  that are applicable in RWSN directly or upon modification. We also present a detailed survey of the limited set of article on the field of RWSN. 

% The rest of this chapter is organized as follows. 

% However, there are some major obstacles for full-fledged practical realization of this integrated field e.g. proper communication, localization, co-ordination, increased control overhead, the requirement of sophisticated and instantaneous control mechanism, and so on. 
% But we are mainly interested in the problems related to the networking aspects of this new promising field such as proper routing, connectivity maintenance, RSS mapping and localization. 
% In the rest of this paper, We present a survey on the such problems of interest in networking among robots, state-of-the art solutions to these problems. We also discuss the state-of-the-art communication models and protocols used in RWSN.  %The rest of the manuscript have been organized as follows.

\section{What is a RWSN?}
\label{sec:rwsn}
In this section, we illustrate the field of RWSN in details. 
Here, we address some core and important questions related to RWSN:
What is an RWSN? What kind of research works are classified as RWSN related works?
% Is any research on collaborative robotics or WSN can also be classified as RWSN related research? 
%In this section, we answer those questions with sufficient examples.
\begin{figure}
\centering
\includegraphics[width=0.8\linewidth]{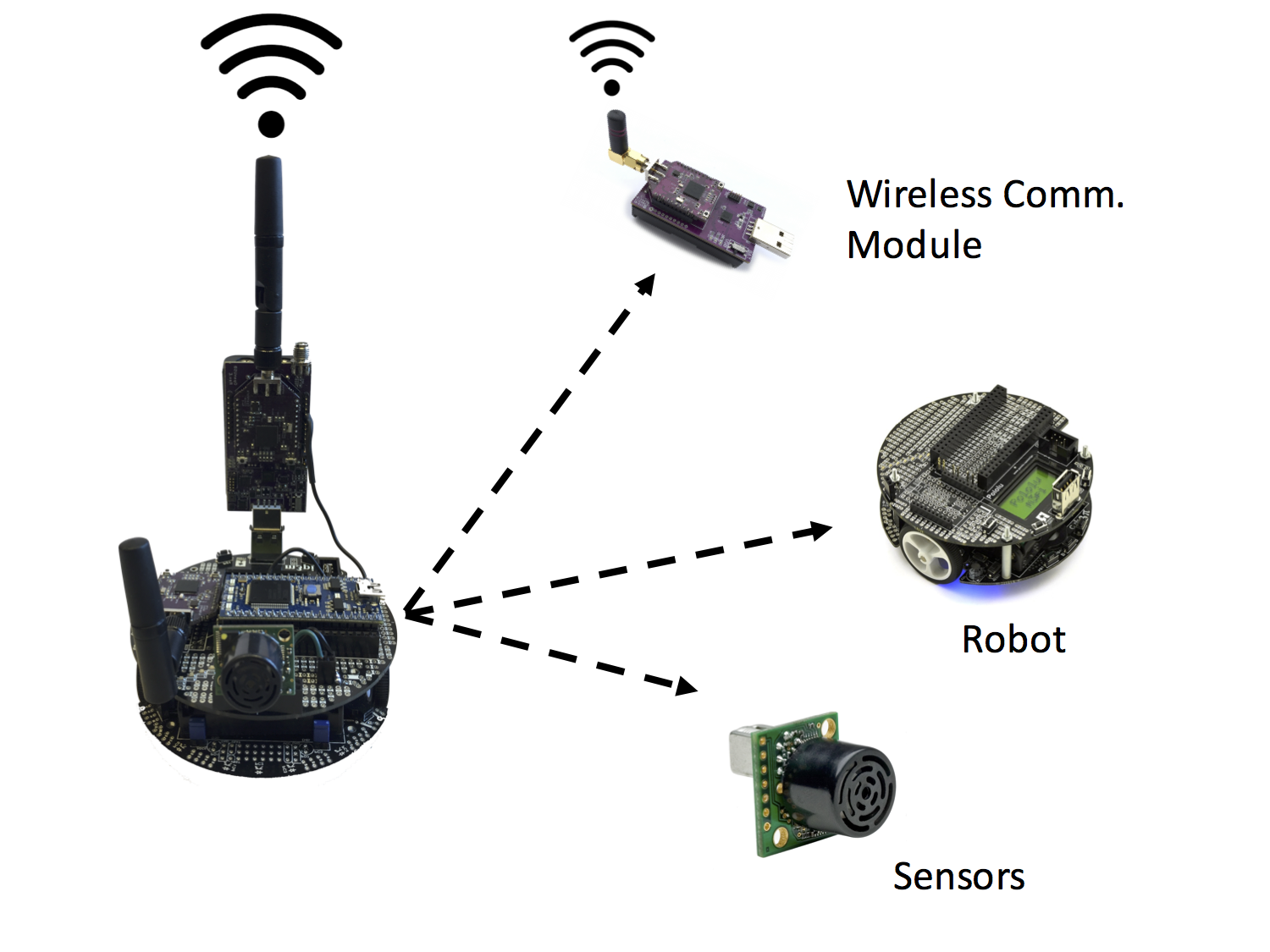}
\caption{Illustration of a Robotic Wireless Sensors}
\label{fig:robo_router}
\end{figure}

We define an RWSN as a wireless network that includes a set of robotic nodes with controlled mobility and a set of nodes equipped with sensors; whereas all nodes have wireless communication capabilities.
Ideally, each node of a robotic sensor network should have controlled mobility, a set of sensors, and wireless communication capabilities (as illustrated in Figure~\ref{fig:robo_router}).
We refer to such nodes (devices) as \textbf{``Robotic Wireless Sensors (RWS)''}.
Nonetheless, an RWSN can also have some nodes with just sensing and wireless communication capabilities but without controlled mobilities.
We refer to such nodes (by following traditional terminology) as \textbf{``Wireless Sensors''}.
Note that, every node of an RWSN must have wireless communication capabilities according to our definition.
Moreover, an RWSN is typically expected to be able to fulfill or guarantee certain communication performance requirements enforced by the application contexts such as minimum achievable bit error rate (BER) in every link of the network.

\begin{figure}
\centering
\includegraphics[width=0.9\linewidth]{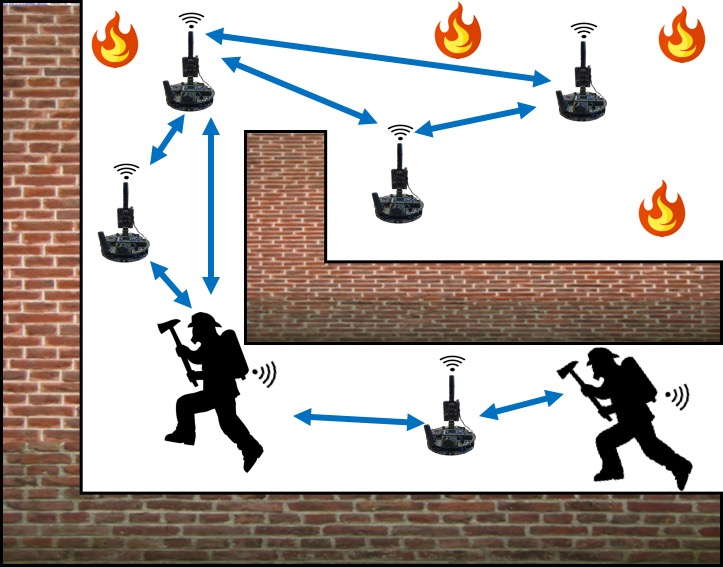}
\caption{Illustration of Robotic Sensor where a group of five robots are sensing the environment around the firefighters to guide them in firefighting while also providing connectivity}
\label{fig:robo_sense}
\end{figure}

To answer the second question, the existing research works related to RWSN can be subdivided into two broader genres. 
The first genre focuses on generic multi-robot sensing systems with \rev{realistic communication channels (i.e., include the effects of fading, shadowing etc.) }between the robots.
\emph{To clarify, these are mostly the existing works in the robotics literature on multi-robot systems but with practical wireless communication and networking models.}
One application context of such a RWSN is in robot assisted fire-fighting where the robots are tasked to sense the unknown environments inside rubble to help and guide the firefighters.
Now, if the robots are not able to maintain a good connectivity among themselves or to a mission control station, the whole mission is voided.
Refer to Figure~\ref{fig:robo_sense} for an illustration of such contexts where a group of robots are sensing an unknown environment to guide the human movements.
In Figure~\ref{fig:robo_sense}, the network consisting of five robots and two firefighters needs to be connected all the time and also needs to have properties such as reliability and lower packet delays.
Thus, we need a class of multi-objective motion control that will optimize the sensing and exploration task performance, and will also ensure the connectivity and the performance of the network. 
Some of the main identifiable challenges in this genre of works are: connectivity maintenance, efficient routing to reduce end to end delay of packets, and multi-objective motion control and optimization.
%\emph{Again, there exist a significant amount of works related to these classic problems, which does not include realistic communication model.
%The communications between nodes in such environments are extremely dynamic (fading, shadowing~\cite{rappaport1996wireless}) in nature due to cluttered and constantly changing environments.
%Thus, the real characteristics of the communication links are far deviant from a simple disk model characteristics.}
%For completeness, we do not completely omit such relevant works and rather point out the missing elements.
 % that includes the network performance optimization and the allocated actual task (such as sensing) performance optimization.

\begin{figure}
\centering
\includegraphics[width=0.9\linewidth]{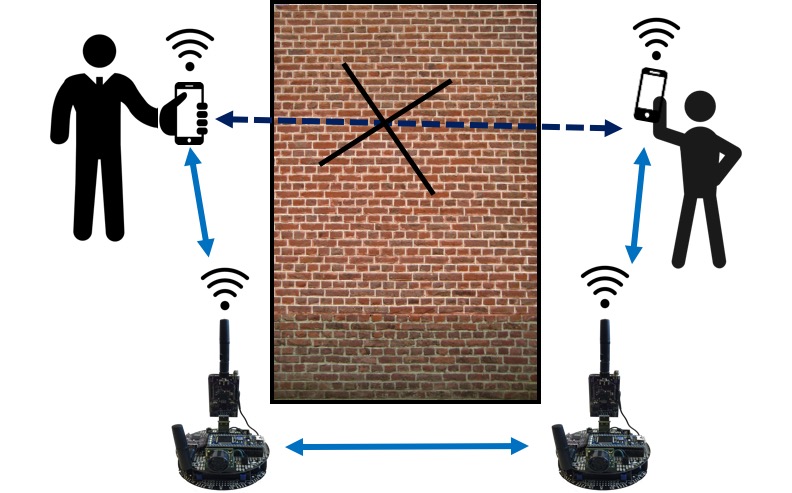}
\caption{Illustration of robotic routers where the two humans are not able to communicate directly due  presence of a wall or some other blocking object}
\label{fig:robo_route}
\end{figure}

The second genre of works focus on the application of RWS to create and support a temporary communication backbone between a set of communicating entities.
In these contexts, we sometime use the terms \textbf{``robotic router''} and \textbf{``robotic wireless sensor''} synonymously, to put emphasis on the communication and routing goals. 
The main theme of these works is to exploit the controlled mobility of the robotic routers to perform sensing and communication tasks.
\emph{Note that, there exists a vast literature on multi-agent systems in robotics and control community that apply simple disk models for communication modeling and, subsequently, apply graph theory to solve different known problems such as connectivity and relay/repeater node placements. 
In order to be directly included in the RWSN literature, these existing works need to include \rev{the effects of fading and shadowing in the communication models} which is likely to significantly increase the complexity of the problems as well as the solutions.}
An example of the second genre of works is in the application of RWS in setting up a temporary communication backbone.
While sensing is still involved for the robotic router placement optimization and adaptation, the main purpose of the system is to support communication, not sensing.
% Let us assume that there exist a set of communication endpoint pairs that want to communicate with each other but unable due to absence of a communication path.
% In such contexts, a group of robotic routers can form a communication relay path between two endpoints to relay message from one end to another. 
In Figure~\ref{fig:robo_route}, we present an example illustration where a set of two robotic routers form a communication relay path between two humans (e.g., two fire-fighters) who are unable to communicate directly.
Some of the main challenges in this genre of RWSN research are: link performance guarantee (in terms of Signal to Interference plus Noise Ratio, SINR, or Bit Error Rate, BER), optimized robotic router placements and movements in a dynamic network,
non-linear control dynamics due to inclusion of network performance metrics into control loop, and localization.  
A special case of this would be robot assisted static relay deployments, where the robots act as carriers of static relay nodes and smartly place/deploy them to form a communication path/backbone. 

Next, we identify a set of system components and algorithms required in an RWSN as follows. 
All these pieces are individual research problems themselves and thus require separate attentions.
%we identify the following types of works (but not limited to) to be categorized as RWSN related works.
\begin{itemize}
\item \textbf{RSSI Models, Measurements, and RF mapping:} 
In an RWSN, it is important to estimate and monitor the quality of the communication links between the nodes (in terms of Bit Error Rate (BER), Signal to Noise plus Interference Ratio (SINR) etc.) in order to satisfy the communication related requirements (Note that, RF based communication is standard mode of communication in RWSN for obvious reasons).
For practicality, these estimations must be either partly or fully based on \rev{online RF sensing such as temporal RSSI measurements in a deployment.}
Moreover, in some application contexts of RWSN, the sole goal of a robotic sensor network can be to sense and formulate an RF map of an environment to be processed or exploited later on.  
We present a survey of such RF mapping and modeling related works in Section~\ref{sec:rf_sensing}.

\item \textbf{Routing Protocols:} Similar to any wireless sensor networks, routing and data collection is crucial in an RWSN.
The concept of RWSN have opened up the door to a new class of routing protocols that incorporates the controlled mobility of the nodes in the routing decisions for more effective communication.
Moreover, end to end delay reduction and reliability improvement have become of prime interests.
A brief survey of existing RWSN related works on routing protocols is presented in Section~\ref{sec:routing}.

\item \textbf{Connectivity Maintenance:} 
In any collaborative network of robots, it is important to maintain a steady communication path (direct or multi-hop) between any pair of nodes in the network for an effective operation.
This problem, traditionally referred to as connectivity maintenance problem, in very well studied by the robotics research community. 
%In most of these connectivity maintenance related works, the main focus is on the existence of a communication path but not on the quality of the communication path.
%In general, any works on multi robot system or swarm robotics that focus on wireless link and end-to-end path quality maintenance (in terms of BER or SNR or SINR) are classified as RWSN related works.
A survey of such state-of-the-arts is presented in Section~\ref{sec:connectivity}.

\item \textbf{Communication Aware Robot Positioning and Movement Control:} 
As mention earlier, one of the application contexts of RWSN is in supporting temporary communication backbones.
\rev{
The most important research question in such contexts is to devise a control system that adapts the positions of the robotic routers throughout the period of deployment to optimize the network performance while optimizing the movements as well.
Therefore, the main goal of these class of work is continuous joint optimization of the robotic movements and the wireless network's performance.
Moreover, the router placement controller should also be able to support network dynamics such as node failures and change in the set of communication endpoints.}
Another important application context of RWSN systems is distributed coordinated sensing using multiple robots.
In such sensing contexts, the robotic sensing agents should be able to optimally sense the region of interest and route the sensed data to other nodes or a command center.
This also require careful communication-channel-dynamics-aware positioning of the robotic sensors.
%Thus, we require a modular control system architecture for the robotic nodes that integrates network performance goal based decisions with the control decisions.
%In such control system we need to optimize the movement and network performance.
We present a summary of such communication aware robotic router/sensor positioning works in Section~\ref{sec:robo_router}.

\item \textbf{Localization:} Localization is a well known problem in the field of WSN as well as robotics. 
Thus, it is quite intuitive to be included in the field of RWSN. 
Moreover, the field of RWSN sometimes requires techniques for robots to follow each other or maintain proximity to each other. \rev{For that high accuracy relative localization is more important than absolute localization.}
%Therefore, relative localization racking and relative position control is also an area of interest.
We present a brief summary of such localization related works in Section~\ref{sec:local}.

%\item \textbf{Unified System Architecture:} 
%Another key challenges in the field of RWSN is to develop a modular control system architecture for the robotic nodes.
%In such control system we need to optimize the movement and network performance.
%Unlike the previous five challenges, this is more focused on the system in entirety that incorporate communication objectives in the movement control.

\end{itemize}

% ---one more potential subfield is robot based data collection.

\section{RWSN Systems Components: }
\label{sec:soa}
In this section, we present a categorical survey of all state-of-the-arts in the field of RWSN.
The works are classified according to the problem addressed. 
%Most of the works on the communication aspect of RWSN can be organized into five groups. 
% The first, that we classify as \emph{RSS models, measurement and RF mapping}, deals with mathematical modeling of the wireless signal strength variation to over region covered by the group of robots to capture effects of fading, shadowing, path loss and interference.
% The second, that we classify as \emph{routing} related works, are mainly the routing protocols that are developed will sole focus on RWSN.
% The third, that we refer to as \emph{connectivity maintenance protocols}, deals with the difficult task of maintaining connectivity while each robot is performing the tasks assigned to it. 
% The fourth, that we refer to as \emph{robotic router and relay placement}, focuses on using robots to form communication bridges between sender-receiver pairs in the absence of any fixed infrastructures.
% The fifth and last, that we refer to as \emph{network \& RF based localization}, focuses on localization of robot nodes based on RSS maps and wireless signal strengths from some \emph{seed nodes}.

\subsection{ RSS Models, Measurements, and RF Mapping}
\label{sec:rf_sensing}
Radio Signal Strength (RSS) variation over a spacial domain greatly impacts the wireless communication properties, such as power decay and packet loss, between two nodes. 
Different properties of a physical environment, such as obstacles and propagation medium, affect radio signal propagation in different way,
thereby, causing fading, shadowing, interference, and path loss effects~\cite{rappaport1996wireless}. 
All these should be taken into consideration (via proper communication channel models) for proper selection of radio transmission and reception parameters to improve the communication quality in a RWSN.
While there exists a range of standard communication channel models in the literature, such as simple path loss model and log normal fading model~\cite{rappaport1996wireless},
the applicability as well the model parameters' values depend on the actual deployment environments.
For example, according to a log normal fading model~\cite{rappaport1996wireless}, the received power is calculated as follows.
\begin{equation}
\label{eqn:power}
\begin{split}
        &P_{r,dBm}=P_{t,dBm}+G_{dB}-L_{ref}-10\eta \log_{10} \frac{d(t)}{d_{ref}} + \psi\\
        &P_{r,dBm}^{ref}=P_{t,dBm}+G_{dB}-L_{ref} + \psi \\
        % &\psi= \psi_{sh,dB} +\psi_{MP,dB}\\
        % &\implies \frac{d(t)}{d_{ref}}=10^{\frac{\left(P_{t,dBm} +G_{dB} -P_{r,dBm}  -L_{ref} + \psi_{SH,dB} +\psi_{MP,dB} +noise_{dB}\right) }{10\cdot \eta}}\\
\end{split}
\end{equation}
where $P_{r,dBm}$ is the received power in dBm, $P_{t,dBm}$ is the transmitter power in dBm, $G_{dB}$ is the gain in dB,
$L_{ref}$ is the loss at the reference distance $d_{ref}$, $\eta$ is the path loss exponent,
$d(t)$ is the distance between the transmitter and receiver,
$\psi \sim \mathcal{N}(0,\sigma^2)$ is the random shadowing and multipath fading noise which is log normal with variance $\sigma^2$,
and $P_{r,dBm}^{ref}$ is the received power at reference distance ($d_{ref}$). 
While some of the variables such as $P_{t,dBm}$, $d_{ref}$, and $G_{dB}$ are known or can be measured, the values of other variables such as $\eta$, $L_{ref}$, and $\sigma$ are dependent upon the deployment environment.
Thus, it is important to identify or estimate the proper values of such parameters in the deployment environment in order to estimate the $P_{r,dBm}$. 
% Simultaneously, mapping of RSS over the region of interest requires proper modeling of the channels. 
Moreover, RSS models and maps are very important and useful database that can be used for a range of purposes such as localization of nodes based on received signal strength~\cite{liu2007survey,sun2005signal,guvenc2009survey,amundson2009survey}, mapping of an unknown terrain~\cite{Mostofi_Sen_ACC09}, and identifying obstacles~\cite{GonzalezGhaffarkhahMostofi13}.
The communication model based estimations of the RF signal strengths in the future and unvisited locations are also very important for maintaining connectivity among mobile nodes and for optimizing the network performance. 
% Cooperative sensing and mapping is also a very important aspect in robo-cup \cite{dietl2001cooperative}.
In this section, we provide a brief overview of the state-of-the-art RSS modeling and mapping techniques that are applicable in a RWSN.
%Note that, the communication models from standard WSN are not considered in this chapter as they are mostly invalid in the context of RWSN.
% In the context of mapping and modeling RSSI or RF channels, researchers tends to opt for two different approaches. 
% In the first kind of approaches, a set of robots are used to first scan the environment to identify the channel properties
% followed by exploiting those precalculated properties for communication modeling.

\textbf{What is already out there?}
In the context of mapping and modeling of the RF channels, the most common and practical class of approaches is to deploy the network of robots with a initial model of the communication channel.
Then, the robots continually or periodically collect RF samples to update the communication model parameters in an online fashion.
One can also opt for an offline modeling where the robots collect a set of RF samples over the region of interest followed by post-processing of all the samples to estimate the communication channel properties.
Nonetheless, the later class of methods is unrealistic and of little interest to us as it can not cope with temporal changes in the communication channel properties, which is a quite common phenomenon.
% \emph{Thus, We will mostly focus on the first class of works in this chapter.}
% somewhat online modelling where the network of robots are deployed measures RSSI or RF signal properties over time and update th
% Among the few researchers in the field of RSS modeling and mapping, the works of Yan Mostofi \et are mentionable.

One of the key challenges of online RF mapping is in the sparsity of RF samples. 
Mostofi \et \cite{gonzalez2012comprehensive} have done significant research in exploiting this sparsity to their advantage for RF channel modeling in networked robot systems.
In \cite{mostofi2008compressed}, Mostofi and Sen presented a technique of RSS mapping by exploiting sparse representation of the communication channel in frequency domain. 
They demonstrated how one can reconstruct the original signal using only a small number of sensing measurements by employing the theory of \emph{compressive sampling}~\cite{candes2006compressive}. 
Later, they utilized the compressive sampling based reconstruction of the signal in the domain of cooperative mapping~\cite{Mostofi_Sen_ACC09} to build a map of the region of interest.   
Mostofi \et~\cite{mostofi2009characterization} further presented an overview of the characterization and modeling of wireless channels for a RWSN.  
In their works, all three major dynamics in a physical wireless communication channel (small-scale fading, large-scale fading, and path loss) are considered~\cite{mostofi2011compressive,malmirchegini2012spatial,mostofi2010estimation}. 
In \cite{mostofi2010estimation}, Mostofi, Malmirchegini, and Ghaffarkhah also presented techniques for channel predictions in future locations of robots based on the compressed sensing based channel models.    
% Mostofi~\cite{mostofi2011compressive} also proposed a technique to map the spatial variation of RSS over a field of interest using a group of mobile cooperative units/robots based on only wireless measurements and communication signal strength  . 
% Mostofi \et also introduced a concept of noninvasive mapping that allows the vehicles to map the obstacles without sensing them directly.
Since all these are completely based on wireless measurements, multi-path fading has a great influence over the results. 
Later on, Yan and Mostofi~\cite{yan2013co} presented a combined framework for the optimization for motion planning and communication planning.
The concept of compressive sensing based signal reconstruction and material dependent RF propagation properties are also employed for RF based imaging and mapping of cluttered objects/regions~\cite{Mostofi_TMC12,GonzalezGhaffarkhahMostofi13,TVT15}.
In these works, RF signal propagation properties (mostly attenuations) of the communication path between pairs of moving robots are used to estimate a structural map of an obstacle or a cluttered region.
Hsieh, Kumar, and Camillo~\cite{hsieh2004constructing} have also demonstrated an RSS mapping technique using a group of robots in an urban environment. 
The goal was to learn the environment's communication characteristics to generate a connectivity graph.
Later on, they used this model for connectivity maintenance problem in a network of robots \cite{hsieh2008maintaining}.
% Thus, this work falls under the second class of mapping algorithms.
In \cite{fink2010online}, Fink and Kumar presented another mapping technique using multiple robots. 
Their goal was to use the mapping for localization of an unknown source using a Gaussian process based maximum likelihood detection.
They further extended the Gaussian process based channel model to guarantee a minimum stochastic end-to-end data rate in a network of robots~\cite{fink2013robust}.
% They also modeled the packet loss. 
Signal attenuation factors due to presence of obstacles are taken into account in the work of Wang, Krishnamachari, and Ayanian~\cite{wang2015optimism}.
Ghosh and Krishnamachari~\cite{GhoshK16a} have proposed a log normal model based stochastic bound on interference power and SINR for any RWSN.
The works presented in~\cite{dagefu2015short,boillot2015large,qaraqe2013statistical} are also related to this context.

\textbf{What would be the potential future research directions?}
To our knowledge, a generic model for interference and SINR estimation is missing in the current literature.
Interference modeling is a key to develop a more realistic model of wireless channels.
Interference should also be taken into account for robotic router placements contexts, where the main goal of the network is to guarantee certain communication performance qualities.
In our opinion, this should be a major focus of future research in this topic.
Moreover, the effects of channel access protocols, such as Carrier Sense Multiple Access (CSMA)~\cite{rappaport1996wireless}, needs to be taken into account in the interference and SINR models.
The existing signal strength models could be extended to achieve this goal.
Nonetheless, to our knowledge, this has remained an unexplored problem in the context of a RWSN. 
Another future research direction would be to perform an extensive set of measurement experiments in \rev{real (rather than simulated)} mines, undergrounds, or fire fighting environments.
This data can be used to identify the RF properties in such environments that can be exploited to the advantage of RWSN system design.

\begin{table}[!ht]
\centering
\caption{Summary of RSS Models, Measurements and RF Mappings Related Works}
\begin{tabular}{ |p{2cm}|p{9cm}| } 
\hline 
\multirow{2}{2cm}{Algorithm Classes}
            & $\bullet$ \textbf{Online:} Modeling during deployment~\cite{gonzalez2012comprehensive,mostofi2008compressed,Mostofi_Sen_ACC09,mostofi2009characterization,mostofi2010estimation,yan2013co} \\
            & $\bullet$ \textbf{Offline:} Modeling before deployment~\cite{hsieh2004constructing,hsieh2008maintaining,GhoshK16a} \\ \hline
\multirow{3}{2cm}{Challenges}
            & $\bullet$ Sparse Sampling~\cite{Mostofi_Sen_ACC09,gonzalez2012comprehensive} \\
            & $\bullet$ Future Location's Signal Prediction~\cite{mostofi2010estimation,yan2013co} \\
            & $\bullet$ Temporal Dynamics~\cite{mostofi2010estimation} \\ \hline
\multirow{4}{2cm}{Available Theoretical Tools}
            & $\bullet$ Compressive Sampling~\cite{candes2006compressive} \\
            & $\bullet$ Gaussian Process~\cite{fink2010online,GhoshK16a} \\
            & $\bullet$ Fading Models~\cite{rappaport1996wireless} \\
            & $\bullet$ Path Loss Models~\cite{rappaport1996wireless} \\ \hline
\multirow{3}{2cm}{Potential Future Directions}
            & $\bullet$ Interference and SINR Models \\
            & $\bullet$ Account for Channel Access Methods like CSMA \\
            & $\bullet$ Real World Data Collection and Analysis \\ \hline
\end{tabular}
\end{table}

\label{sec:rf_mapp}

\subsection{Routing}
\label{sec:routing}
Routing in an RWSN can be considered same as in Mobile Adhoc Networks (MANET) but with an extra advantage of controllability. 
In MANET, there are mainly two types of popular routing algorithms called reactive and proactive techniques.
Among the reactive techniques, Ad-hoc On demand Distance Vector (AODV) \cite{das2003ad,chakeres2004aodv} and Dynamic Source Routing (DSR) \cite{deering1998internet,johnson2007dynamic} are the most popular ones. 
On the other hand, in the class of proactive techniques, Optimized Link State Routing (OLSR) \cite{clausen2003optimized} and B.A.T.M.A.N. \cite{johnson2008simple} are the popular ones.
{While any of these algorithms can be used for an RWSN, they do not take advantage of the extra feature in RWSN: controlled mobility.
Ideally, the controlled mobility aspect should also be taken into account for optimized routing decisions.
Nonetheless, it remained to be one of the less explored areas in RWSN.
Moreover, a lower end-to-end delay and higher reliability in packet routing (mostly control packets) are two important and required aspects in an RWSN. 
Delayed or missing packets can result in an improper collaborative movement control and task completion in an RWSN.}
To this extent, some researchers have modified existing routing solutions to adapt in a robotic network and proposed completely new routing solutions as well. 
In this section, we present a brief survey of the state-of-art routing techniques in RWSN that are developed or modified with sole focus on robotics.

\textbf{MRSR, MRDV and MRMM: }
In \cite{das2007mobility}, Das \et presented three routing protocols based on traditional mobile ad-hoc protocols, such as DSR~\cite{deering1998internet} and AODV~\cite{chakeres2004aodv}, for routing in a network of mobile robots.
A brief description of each of these algorithms are as follows.

\begin{itemize}
	\item \emph{Mobile Robot Source Routing (MRSR):} It is a unicast routing algorithm based on Dynamic Source Routing (DSR) \cite{johnson1996dynamic}.
    MRSR incorporates three mechanisms: \emph{route discovery}, \emph{route construction}, and \emph{route maintenance}.
    In the runtime of \emph{route discovery} phase, each robot along the pathway of \emph{route reply} message encodes its mobility information into the route reply packet.
    During \emph{route construction}, MRSR exploits graph cache that contains the topological information of the network. The \emph{route maintenance} phase is similar to the maintenance method applied in DSR.
	\item \emph{Mobile Robot Distance Vector (MRDV):} This is also a unicast routing algorithm based on the well known AODV \cite{perkins1999ad} routing protocol.
    MRDV protocol adopts AODV features such as the on-demand behavior and hop-by-hop destination sequence number.
    Nevertheless, unlike MRSR, MRDV explores only one route that may not have the longest lifetime among all possible routes, thereby, resulting in high probability of route errors.
	\item \emph{Mobile Robot Mesh Multicast (MRMM):} This is a multicast protocol for mobile robot networks based on ODMRP (On Demand Multicast Routing Protocol) \cite{lee2002demand}  for MANETs.  
    % Unlike original ODMRP protocol, MRMM includes some application specific features of robots.
\end{itemize}

\textbf{Adaptive Energy Efficient Routing Protocol (AER): } 
This protocol was proposed by Abishek \et \cite{abishek2012aer} to achieve optimal control strategy for performing surveillance using a network of flying robots.  
AER protocol is also subdivided into three phases (similar to DSR~\cite{johnson1996dynamic}): \emph{route discovery}, \emph{route maintenance}, and \emph{route failure handling}. 
The residual energy levels and signal strengths at the neighboring nodes are the main route determining factors in this protocol.
To model them, the authors defined two decision parameters: T (attribute value of the neighbor) and C (cost function). The best value of T decides the forwarding node. 
The nodes that are neither selected for message forwarding nor have sufficient energy, are switched to the sleep state to minimize energy consumptions. 
% This protocol also uses a new data format of control message, referred to as FAGENT, that includes a unique sequence number, hop length field, destination address, and a list of all intermediate nodes in its way to the destination.

\textbf{ACTor based Robots and Equipments Synthetic System (ACTRESS): } 
In \cite{matsumoto1990communication}, Matsumoto \et proposed a robotic platform called ACTRESS that consists of robotic elements referred to as \emph{robotors}.
They also proposed a routing protocol exclusively for that platform. 
The messages are classified into two different classes: messages to establish/relinquish a communication link, and messages for control and rest of the purposes. 
The first kind of messages use traditional communication protocols such as TCP/IP. 
The second type of messages use their proposed special protocol to establish logical links, allocate tasks, and control cooperative motions.
For this purpose, the authors have introduced four levels of messages: Physical level, Procedural level, Knowledge level, and Concept level.
The common part of all four types of messages is referred to as the \emph{message protocol core}, which is used for: negotiation, inquiry, offer, announcement and synchronization. 

% \begin{itemize}
% 	\item \emph{Physical level:} This group of messages are related to physical information such as position, velocity, or force. 
% 	\item \emph{Procedural level: } These type of messages are used to command \emph{robotors} to operate with its specified parameters, if any. 
% 	\item \emph{Knowledge level:} This level handles the condition status that are extracted from knowledge base or obtained from other \emph{robotors}. 
% 	\item \emph{Concept level: } These messages deal with concepts such as intention or target.
% \end{itemize}

\textbf{WNet:} 
Tiderko \et \cite{tiderko2008rose} proposed a new multi-cast communication technique called WNet that is based on the well-known Optimized Link State Routing(OSLR) protocol~\cite{clausen2003optimized}. 
% This algorithm was proposed mainly for a new mobile robot programming framework, referred to as the Robot Services (RoSe) network.
% In this framework, a quality-based routing mechanism monitors the fundamental attributes of all wireless links.
% By exploiting the monitored data, a communication path is chosen by WNet. 
Similar to OLSR, WNet uses HELLO and Topology Control (TC) management frames to create and update the network topology graph stored in each robot node. 
However, the packet frames are integrated with some additional information of link attributes that is used for link quality estimation. 
Next, the Dijkstra algorithm \cite{dijkstra1959note} is applied to the topology graphs to determine the routing paths.

\textbf{Steward Assisted Routing (StAR):}
 In \cite{weitzenfeld2006multi}, Weitzenfeld \et presented a new routing algorithm called StAR that deals with mobility and interference in an RWSN. 
 The objective of this protocol is to nominate, for each connected partition, a \emph{``steward''} for each destination. 
 These \emph{stewards} are noting but next hop robots toward destination  that can store the data until a route to the destination is available. 
 The message routing of StAR is based on a combination of global contact information and local route maintenance. 
 Also periodic broadcast messages with unique source identifiers are sent containing topological location of the active destination.
 % This sequence number is an identifier of the broadcast source. 
 At the beginning of this process, each node selects itself as the steward and then progressively changes the local steward based on advertisements from the neighbors.
 StAR uses a sequence number to maintain the freshness of information, similar to AODV protocol. %
%The stewards ranked according to the most recently heard sequence number for a destination. If two nodes have same destination sequence number, route length is used for tie-breaker. 

\textbf{Optimal Hop Count Routing (OHCR) and Minimum Power over Progress Routing (MPoPR):} 
{Hai, Amiya, and Ivan \cite{liu2007localized} are among the few researchers who leveraged controlled mobility of the robotic routers to assist in wireless data transmission among fixed nodes. 
This method is divided into two parts.}
The first, which they refer to as Optimal Hop Count Routing (OHCP), computes the optimal number of hops and optimal distances of adjacent nodes on the route.
Each node identifies its closest node by comparing the respective neighbors' distances with the optimal distance.  
If a node cannot find any such neighboring node, it sends back a route failure message to the source. 
Otherwise, the second part of the routing, which the authors refer to as Minimum Power over Progress Routing (MPoPR), uses greedy routing on the results obtained from OHCP to minimize the total transmission power.  

\textbf{Synchronized QoS routing:} 
In \cite{sugiyama2006qos}, Sugiyama, Tsujioka, and Murata presented a QoS routing technique for a robotic network that is based on the Qos Routing in Ad-hoc network \cite{lin1999qos} and DSDV routing protocol \cite{perkins1994highly}. 
In this method, they used the concept of Virtual Circuits to reserve a specified bandwidth. 
It is the job of the sender to reestablish the circuit in case of a broken connection due to topology changes. 
This method also includes a methodology for accelerating the transmissions of the control packets.

\textbf{Topology Broadcast based on Reverse-Path Forwarding (TBRPF):} 
In the CENTIBOT project \cite{konolige2003large}, Konolige \et used a proactive MANET technique called Topology Broadcast based on Reverse-Path Forwarding (TBRPF) to deal with multi-hop routing in dynamic robotic network. 
This link-state routing protocol was originally proposed by Bellur and Ogiel \cite{bellur1999reliable,ogier2004topology}. 
In this algorithm each node maintains a partial source tree and report part of this tree to its neighbor.  
To deal with mobility it uses a combination of periodic and differential updates. 

\textbf{B.A.T.Mobile:}
Sliwa \et \cite{sliwa2016bat} have also proposed a mobility aware routing protocol called B.A.T.Mobile which  builds upon the well-known B.A.T.M.A.N routing~\cite{johnson2008simple} protocol for MANET. 
This algorithm relies on a future position estimation module for the next hops that uses the current and past position related information as well as the knowledge of the mobility algorithms of the users. The estimation module is further used to rank the neighbors and estimate their lifetime. The neighbor rankings are used to change the route in a proactive manner for end-end to data transfer. 

\textbf{Other Methods:} 
There also exist some methods that have the potential to be used in an RWSN after few modifications.
Among such methods, the geographic routing algorithms and encounter based routing algorithms are mentionable. 
%A multipath version of Dynamic source routing is presented in \cite{nasipuri1999demand}. Their technique is based on disjoint paths. They also presented an analytic model demonstrating the advantages of multipath techniques over signgle path techniques.
%	
%	
%\subsubsection{Position-Based Routing/ Geographic Routing:}
 Greedy Perimeter Stateless Routing (GPSR) \cite{karp2000gpsr} is an example of geographic routing protocols that uses router positions and packet destinations for making forwarding decisions. 
 % This algorithm uses information from immediate neighbor only. 
 If the locations of all nodes in the network are known, this algorithm can be used in RWSN. 
 However, this approach faces many problems in mobile wireless networks. 
 In \cite{son2004effect} Son, Helmy and Krishnamachari identified two problems due to mobility in geographic routing, particularly in GPSR, called LLNK and LOOP. 
 They also presented two solutions: neighbor location prediction (NLP) and destination location prediction (DLP); to solve those problems.
 Rao \et \cite{rao2008gpsr} also identified some issues with GPSR  and proposed a lifetime timer based solution.
 In \cite{mauve2003position}, Mauve \et presented a generalized multicast version of GPSR like geographic routing. 
 There are many other works on position based routing \cite{lochert2003routing,karp2001challenges}.
 For a more detailed and complete overview on position based routing algorithms, an interested reader is referred to \cite{mauve2001survey}.

Encounter based routing is another relevant group of routing, mainly used in delay tolerant networks (DTN). 
In general, DTN routing protocols are divided into two categories: forwarding-based or replication-based. 
Forwarding-based protocols use only one copy of the message in the entire network while the replication based technique uses multiple copies of the message.  
Replica based protocols are also subdivided to two categories: quota-based and flooding-based.
Flooding is the most simple and inefficient technique. 
Balasubhamanian, Levine, and Venkataramani presented a flooding based technique of replication routing in DTN \cite{balasubramanian2010replication,balasubramanian2007dtn}, modeling it as a resource allocation problem. 
Another flooding based technique is presented in \cite{burgess2006maxprop}, called Maxprop. 
Spyropoulos, Psounis and Raghavendra presented two quota based replication routing techniques for DTN called Spray and Wait \cite{spyropoulos2005spray}, and Spray and Focus \cite{spyropoulos2007spray}. 
There are many other papers on DTN routing \cite{li2008dtn,fall2008dtn,fall2003delay}.
Although, this  group of techniques are not directly related, they can be modified to develop very efficient routing for RWSN. 

%Later on many people have proposed some modified techniques of Spray and Wait routing \cite{wang2010dynamic,huang2011spray}.
% In \cite{nelson2009encounter}, Nelson, Bakht and Kravets presented a quota-based DTN routing technique called Encounter-based Routing (EBR). They used a prediction model for future encounters based on the past data. The NECTAR protocol\cite{de2009nectar} is another example of STN routing algorithm based on the past contacts history. In \cite{sandulescu2008opportunistic}, Sandulescu and Nadjm-Tehrani presented a contact window based opportunistic routing protocol called ORWAR. 

{
The research works related to data collection protocols in WSN community are also of interest.
Among these protocols, a prominent and recent class of queue aware routing algorithms, called Backpressure routing algorithms \cite{tassiulas1992stability,moeller2010routing}, has caught our interest.
The Backpressure routing algorithms and a range of similar algorithms~\cite{dai2008asymptotic,shah2006optimal, naghshvar2012general} are proved to be `throughput optimal', in theory.
One of the most recent Backpressure style routing algorithm is called the Heat Diffusion (HD) routing algorithm~\cite{banirazi2014heat, banirazi2014dirichlet} that has shown to offer a Pareto-optimal trade-off between routing cost and queue congestion (delay).
The Backpressure routing algorithms, including HD algorithm, do not require any explicit path computations.
Instead, the next-hop for each packet depends on queue-differential weights that are functions of the local queue occupancy information and link state information at each node.
There have been several reductions of Backpressure routing to practice in the form of distributed  protocols, pragmatically implemented and empirically evaluated for different types of wireless networks~\cite{moeller2010routing, Martinez11, Alresaini12,BackIP}.
Ghosh \et have also developed a distributed practical version of the HD algorithm called Heat Diffusion collection protocol~\cite{GhoshRBKJ16}.
While these protocols perform effectively in a static WSN, their applicability in an RWSN are yet to be tested.
Since these algorithms do not require any route calculation as well as routing tables, they will require less memory and computation in the resource constrained robotic nodes.
Moreover, one extra advantage of such protocols is the adaptability in a dynamic network due to not relying on a single predetermined path.
Besides these protocols, there exist a number of other prior works on routing and collection protocols for wireless sensor networks, including the Collection Tree Protocol (CTP)~\cite{gnawali2009collection}, Glossy~\cite{ferrari2011efficient}, Dozer~\cite{burri2007dozer}, Low-power Wireless Bus~\cite{ferrari2012low}, and RPL~\cite{RPL}.
These protocols can also be modified for application in RWSN.
The work presented in Glossy~\cite{ferrari2011efficient} is of particular interest due to its simplicity and wide adaptability for high throughputs and low delays.
}

\textbf{What would be the potential future research directions?}
To our knowledge, there exist a significant amount of research on routing related to MANET and WSN that can be applied to an RWSN either directly or after some modifications.
However, a significant focus of future routing algorithms need to be directed towards reducing delays, improving reliability, and incorporating the controlled mobility in the routing decisions.
The emphasis should be on delay and reliability as on-time message delivery among different control system components is the key for a successful and efficient control system.
One example of using the controlled mobility to our advantage is shown in the works of Wang, Gasparri, and Krishnamachari~\cite{wang2013robotic} where the robots ferry messages from a source to sink in a way similar to a postman.
\rev{Another research direction would be to add node movements in routing decisions. For illustration, assume that there exists two possible routing paths and the relatively bad path can be improved considerably by slightly moving the node. Then, the routing decision should include movement into consideration.} 

\begin{table}[!ht]
\caption{Summary of Routing Related Works}
\begin{tabular}{ |p{4.5cm}|p{2cm}|p{4.5cm}| } 
    \hline
    Routing Algorithm Name/Class & References & Comments \\ \hline
    MRSR, MRDV and MRMM & \cite{das2007mobility} & Based on Dynamic Source Routing (DSR) \cite{johnson1996dynamic}, AODV \cite{perkins1999ad}, and On Demand Multicast Routing Protocol \cite{lee2002demand}, respectively\\ \hline
    Adaptive Energy Efficient Routing Protocol(AER) &  \cite{abishek2012aer} & Similar to DSR~\cite{johnson1996dynamic} \\ \hline
    ACTor based Robots and Equipments Synthetic System (ACTRESS) & \cite{matsumoto1990communication} & Main steps are negotiation, inquiry, offer, announcement and synchronization. \\ \hline
    WNet & \cite{tiderko2008rose} & Based on Optimized Link State Routing(OSLR) protocol~\cite{clausen2003optimized} \\ \hline
    Steward Assisted Routing (StAR) & \cite{weitzenfeld2006multi} & It is a hierarchical routing protocol where a group of robots act as "stewards" that can store the data until
a route to the destination is available. \\ \hline
    Optimal Hop Count Routing (OHCR) and Minimum Power over Progress Routing (MPoPR) & \cite{liu2007localized} & Uses the controlled mobility of the robotic routers to assist
in wireless data transmission among fixed nodes.   \\ \hline 
    Synchronized QoS routing & \cite{sugiyama2006qos} &  Based on the Qos Routing in Ad-hoc networks \cite{lin1999qos} and DSDV routing protocol \cite{perkins1994highly} \\ \hline
    Topology Broadcast based on Reverse-Path Forwarding (TBRPF) & \cite{konolige2003large} & It is a link state  routing protocol. Each node maintains a partial tree. \\ \hline
    B.A.T.Mobile & \cite{sliwa2016bat} & Based on B.A.T.M.A.N routing~\cite{johnson2008simple} \\ \hline
    Geographic Routing Algorithms & \cite{karp2000gpsr,son2004effect,rao2008gpsr,mauve2003position,lochert2003routing,karp2001challenges,mauve2001survey} 
    & Employs locations of the nodes for efficient routing in WSN\\ \hline
    Encounter Based Routing &  & \multirow{3}{4.5cm}{Routing algorithms for delay tolerant networks} \\ 
    $\bullet$ Flooding Based & \cite{balasubramanian2010replication,balasubramanian2007dtn,burgess2006maxprop} & \\     
    $\bullet$ Quota-Based & \cite{spyropoulos2005spray,spyropoulos2007spray} & \\  \hline
    Data Collection Routing in WSN & & \multirow{8}{4.5cm}{Data collection routing algorithms are used in WSNs for efficient routing of the sensed data to the data sinks.} \\
    $\bullet$ Backpressure Routing & \cite{tassiulas1992stability,moeller2010routing} & \\
    $\bullet$ Heat Diffusion Routing & \cite{GhoshRBKJ16,banirazi2014heat, banirazi2014dirichlet} & \\
    $\bullet$ Glossy & \cite{ferrari2011efficient} & \\
    $\bullet$ Collection Tree Protocol & \cite{gnawali2009collection} & \\
    $\bullet$ Dozer & \cite{burri2007dozer} & \\
    $\bullet$ Low-power Wireless Bus & \cite{ferrari2012low} & \\
    $\bullet$ RPL & \cite{RPL} & \\ \hline \hline
    \multirow{4}{4.5cm}{Potential Future Directions} & \multicolumn{2}{p{6.5cm}|}{$\bullet$ Modify existing routing protocol to include controlled mobility} \\ 
    % & \multicolumn{2}{p{6.5cm}|}{$\bullet$ Sensor data collection abstraction (e.g., publish-subscribe model) for multiple dissimilar robotic systems} \\ 
    & \multicolumn{2}{p{6.5cm}|}{$\bullet$ Focus more on delay reduction, data transfer guarantee, and energy efficiency} \\ \hline

\end{tabular}
\end{table}

\ \\

%%Connectivity Maintainace Protocol
\subsection{Connectivity Maintenance}
\label{sec:connectivity}

Connectivity maintenance is a well studied and classic problem in the field of
swarm robotics.  In the connectivity maintenance problem, the main goal is to
guarantee the existence of end-to-end paths between every pairs of nodes. The
interaction between pairs of robots is usually encoded by means of a graph, and
the existence of an edge connecting a pair of vertexes represents the fact that
two robots can exchange information either through sensing or communication
capabilities. \emph{Notably, the connectivity of the interaction graph
represents a fundamental theoretical requirement for proving the convergence of
distributed algorithms in a variety of tasks, ranging from distributed
estimation~\cite{Saber:2004,Xiao:2005,Franceschelli:2014} to distributed
coordination and formation control~\cite{Maggiore:2005,Yang:2008,Guo:2014}.}

% In case of distributed network implementation where every node is not directly connected to the base station, connectivity is very crucial. 

% \textcolor{blue}{In the context of robotic networks, where the concept of connectivity is strictly related to the motion of the robots, two possible version of the connectivity maintenance problem can be defined: local connectivity and global connectivity. The local version of the  connectivity maintenance problem  focuses on the preservation of the original set of links of the graph encoding the robot-to-robot interactions in order to ensure its connectedness.  However, preserving each of the links of the interaction graph may significantly constraint the robots mobility, while in general not each link is strictly required to  ensure  the connectivity of the interaction graph.  The  global version of the connectivity maintenance problem focuses on the preservation of the overall graph connectedness, that is  links can be added or removed as long as this does not prevent the interaction graph to remain connected over time.  
% }

\textbf{Traditional Approach:} In the context of robotic networks, where the
connectivity of the interaction graph is strictly related to the motion of the
robots, a fundamental challenge is the design of distributed control algorithms
which can guarantee  that the relative motions of the robots do not result in a
network partitioning, by relying only on local information exchange.   Two
possible versions of the connectivity maintenance problem can be considered:
local connectivity and global connectivity.  The local version of the
connectivity maintenance problem focuses on the preservation of the original set
of links of the graph encoding the pairwise robot-to-robot interactions to
ensure its connectedness. The global version of the connectivity maintenance
problem focuses on the preservation of the overall graph connectedness, i.e.,
links can be added or removed as long as this does not prevent the interaction
graph to remain connected over time. Historically speaking, the local version of
the connectivity problem has been the first version of the problem to be
investigated by the research community. However, it turned out that preserving
each of the links of the interaction graph  significantly constraint the robots
mobility, while, in general, not each link is strictly required to  ensure  the
connectedness of the interaction graph. For this reason, more recently the
research community has focused mostly on the global version of the problem.
Next, we present a brief survey of the available connectivity maintenance protocols that can be used in RWSN.

As already mentioned, connectivity maintenance has been studied extensively in the contexts of distributed robotics and swarm robotics.
Most of the state-of-the-art protocols for connectivity maintenance are based on a graph modeling of the robot-to-robot interactions.
More specifically, let  $\mathcal{G}= \{ \mathcal{V}, \mathcal{E} \}$ be the interaction graph where $\mathcal{V}=\{1, \ldots, n\}$
is the set of robots and $\mathcal{E}$ is the set of edges encoding the interactions between pairs of neighboring robots. 
In particular, the existence of an edge is often related to the spatial proximity between pairs of robots, i.e.,
an edge exists between two robots if the euclidean distance between them is less than a given threshold.

By following this graph-based modeling of a robotics network, a natural metric to measure the network connectedness is
the \emph{algebraic connectivity}.  More specifically, in the context of graph theory,  the \emph{algebraic
connectivity} is defined as the second smallest Eigenvalue, ${\lambda }_2 (\mathcal{L })$, of the graph Laplacian matrix,
$\mathcal{L} $, of the network. In \cite{fiedler1973algebraic,mohar1991laplacian}, it is shown that ${\lambda }_2
(\mathcal{L})$ is a concave function of the Laplacian matrix and represents the network connectivity when $\mathcal{L}$
is a positive definite matrix.  Thus, the connectivity optimization goal becomes simple maximization of the algebraic
connectivity value, ${\lambda }_2 (\mathcal{L })$.  Another way to represent the connectivity is via a powered sum of
adjacency matrix, $\mathcal{A}_{sum}=\sum_{i=0}^{K}{\mathcal{A}^i}$ where $\mathcal{A}$ is the adjacency matrix of the
network graph $\mathcal{G}$.  $\mathcal{A}_{sum}$ basically represents the number of paths up to length $K$ between
every pair of nodes in the graph~\cite{godsil2001algebraic}.  It follows that for a network to be connected, for all
pairs of nodes and $K =n-1$, $\mathcal{A}_{sum}$ has to be positive definite (where $n$ is number of nodes).

Next, we briefly discuss some representative state-of-the-art local and global methods for connectivity maintenance. 
%We find that most of the old works on the connectivity maintenance, more specifically the works before year 2008 are relies on local connectivity thus restricting the motion of the robotic network. Note that, most of the recent connectivity maintenance algorithm, regardless of being local or global, can have decentralized implementation. 
% by some representative 
%Since the calculation of the graph Laplacian and the algebraic connectivity requires the network configuration, global technique seems to be most suitable option~\cite{kim2006maximizing}. 
%Nonetheless, researcher have proved that, global connectivity have many limitations which can only be overcomed by a decentralized implementation.
In~\cite{Dimarogonas:2008hz}, Dimarogonas and Kyriakopoulos presented one illustrative local connectivity maintenance approach using two potential fields in the controller where the nodes try to maintain all the initial links throughout the time.
Notarstefano \et also presented a double integrator disk graph based local approach for connectivity maintenance~\cite{notarstefano2006maintaining,savla2009maintaining}.
In \cite{spanos2004robust}, Spanos \et introduced a concept of geometric connectivity robustness which is basically a function to model and optimize local connectivity. 
Another class of local connectivity related works lies in the context of leader-follower robot architectures where the follower robots try to maintain the connectivity to a designated leader or vice verse.
The work of Yao and Gupta~\cite{yao2009backbone} is relevant in this context. They employed a leader follower control architecture for connectivity by adaptively classifying the nodes into backbone and non-backbone nodes.
Gaustavi \et  \cite{gustavi2010sufficient} have followed similar path by identifying sufficient conditions for connectivity in a leader-follower architecture of mobile nodes. 
On the other hand, there exists a range of global connectivity related works that are proposed over last decade.
One of the earlier global decentralized connectivity maintenance techniques is the super-gradient and orthonormalization based approach by Gennaro and Jadbabaie~\cite{degennaro2006decentralized}. 
Later on, Dimarogonas and  Johansson~\cite{dimarogonas2010bounded} proposed a control strategy using `bounded' inputs. 
Another very effective approach for connectivity maintenance based on decentralized estimation is presented in~\cite{yang2010decentralized}.
There are many extensions to this framework such as the integration of additional (bounded) control terms~\cite{Sabattini:2013} and the saturation of the connectivity control term itself~\cite{Gasparri:2017}. 
Zavlanos \et  also presented a couple of important techniques on the distributed global connectivity control~\cite{zavlanos2013network, zavlanos2008distributed, zavlanos2007potential} along with a compact survey on graph theoretic approaches for connectivity maintenance~\cite{zavlanos2011graph}.  
In \cite{knorn2009framework}, a technique based on dynamics of consensus methods is presented. 
%Cornejo and Lynch proposed an idea of connectivity service to control path planning in \cite{cornejo2008connectivity}.
Schuresko \et \cite{schuresko2009distributed} also presented techniques for connectivity control based on information dissemination algorithm, game theory, and the concept of spanning tree.
A multi-hop information based global connectivity maintenance and swarming technique is introduced by Manfredi~\cite{manfredi2013algorithm}.
Gil, Feldman, and Rus~\cite{gil2012communication} proposed a well-known k-center problem based connectivity maintenance algorithm for an application context where a group of robotic routers provides routing support to a set of robotic clients.
The concept of bounded velocity of the routers and the clients is employed in this work. 
\textcolor{black}{In~\cite{Williams:2015}, the connectivity maintenance problem in multi-robot systems with unicycle kinematics is addressed. In particular, by exploiting techniques from non-smooth analysis,  a
global connectivity maintenance  technique under non-holonomic kinematics is proposed,  which only requires intermittent estimation of algebraic connectivity, and accommodates discontinuous spatial interactions among robots.} 
Most of these global connectivity maintenance methods are built upon the concepts of graph theory and algebraic connectivity.
% \emph{Note that, there exist many other works on local connectivity maintenance.
% We apologize for not able to include all the references, to keep the survey length manageable.}
 
\textbf{Realistic Approach for RWSN:} While all the previous mentioned methods for connectivity are relevant to the field of RWSN, \rev{most of them lack a communication channel model that includes the effect of fading and shadowing observed in a standard wireless channels.}
Rather, most of these methods employ the simple unit disk model for wireless communication links to model the network graph where every pair of nodes are assumed connected if and only if they are located within a communication radius, say $R$, of each other and disconnected otherwise. 
However, in reality, the communication links are very dynamic and unpredictable due to effects like fading and shadowing~\cite{rappaport1996wireless}. 
Therefore, the unit disk model based connectivity maintenance are rather impractical and should be modified.
Moreover, the connectivity maintenance algorithms should use an optimization function that takes into account communication link features such as signal strengths, data rates, realistic communication models, and line of sight maintenance to define connectedness. 
As an example, Mostofi~\cite{mostofi2009decentralized} presented a realistic communication model based decentralized motion planning technique for connectivity maintenance.
A behavioral approach for connectivity that takes into account the locations, measured signal strength and a map based prediction of signal strengths is proposed by Powers and Balch~\cite{powers2007value}. 
Anderson \et presented a line of sight connectivity maintenance technique via a network of relays and clusters of nodes  in \cite{anderson2003maintaining}.
In \cite{tardioli2010enforcing}, a spring-damper model based connectivity maintenance is described.  
% The work of Ning \et \cite{ning2014connectivity} is also mentionable in this context. 
In summary, there exists only a handful of connectivity protocols that incorporate the well-known characteristics of a RF channel such Fading and even fewer are practically implemented and evaluated. 
 
\textbf{What would be the potential future research directions?}  One obvious future direction would be to extend the
theory of traditional unit disk model based connectivity maintenance protocol to include the effects of fading and shadowing.
A modular or hierarchical approach would be ideal in this context where a  graph identification module (by including fading and shadowing effects) and a graph theory based connectivity maintenance module will work independently
but with synergy. Another future direction would be to develop more protocols of second kind and evaluate them
extensively to enrich the literature. Lastly, but most importantly, there is a lack of real-world experiments with a physical RWSN testbed to validate most of the existing theoretical and algorithmic contributions. Thus, a future goal of
connectivity maintenance related research \rev{should be on the development of a cheap, scalable, and easily programmable physical system} and demonstration of the feasibility of the well-known solutions.

\begin{table}[!ht]
\centering
\caption{Summary of Connectivity Maintenance Related Works}
\begin{tabular}{ |p{1.5cm}|p{9.5cm}| } 
 \hline 
 % --- & ---\\ \hline
\multirow{4}{1.5cm}{Importance}
            & $\bullet$ Exchange of Information \\
            & $\bullet$ Proof of Convergence \\
            & $\bullet$ Distributed Coordination and Formation Control \\
            & $\bullet$ So on \\ \hline
\multirow{2}{1.5cm}{Versions}    
            & $\bullet$ Local \cite{Dimarogonas:2008hz,spanos2004robust,yao2009backbone} \\
            & $\bullet$ Global \cite{dimarogonas2010bounded,knorn2009framework,schuresko2009distributed,Williams:2015,gil2012communication} \\ \hline

\multirow{11}{1.5cm}{Traditional Approach}
            & $\bullet$ \textbf{Algebraic Connectivity:} \\
            & $\bullet \bullet$ Second smallest Eigenvalue, ${\lambda }_2 (\mathcal{L })$, of Graph Laplacian $\mathcal{L}$. \\
            & $\bullet \bullet$ Represents Connectivity if $\mathcal{L}$ is positive definite.\\
            & $\bullet \bullet$ Maximize ${\lambda }_2 (\mathcal{L })$ to improve connectivity \\ \cline{2-2}
            & $\bullet$ \textbf{Powered Sum of Adjacency Matrix: }\\
            & $\bullet \bullet$ $\mathcal{A}_{sum}=\sum_{i=0}^{K}{\mathcal{A}^i}$ where $\mathcal{A}$ is the adjacency matrix  \\
            & $\bullet \bullet$ $\mathcal{A}_{sum}$ represents the number of paths up to length $K$ between any pair of nodes in the graph \\ \cline{2-2} \cline{2-2}
            & $\bullet$ \textbf{Issues:} \\ 
            & $\bullet \bullet$ Relies on simple unit disk model for interactions. \\
            & $\bullet \bullet$ Lack realistic communication channel model; effects of fading and shadowing~\cite{rappaport1996wireless} are ignored. \\
            & $\bullet \bullet$ No Focus on the communication link qualities. \\ \hline

\multirow{6}{1.5cm}{Realistic Approach}
            & $\bullet$ \textbf{Features} \\
            & $\bullet \bullet$ Accounts for location, signal strengths, interference, and data rates \cite{mostofi2009decentralized,powers2007value} \\
            & $\bullet \bullet$ Realistic communication channel models~\cite{mostofi2009decentralized}. \\
            & $\bullet \bullet$ Line of sight maintenance between neighbors~\cite{anderson2003maintaining}. \\ \cline{2-2}
            & $\bullet$ \textbf{Examples} \\
            & $\bullet \bullet$ \cite{mostofi2009decentralized,powers2007value,anderson2003maintaining,tardioli2010enforcing,ning2014connectivity} \\ \hline

\multirow{3}{1.5cm}{Potential Future Directions}
            & $\bullet$ Modify existing unit disk model based graph methods of connectivity\\
            & $\bullet$ Develop more efficient algorithm for connectivity with the focus on the realistic link qualities \\
            & $\bullet$ Develop hardware prototypes and test out the algorithm in real world scenarios \\ \hline

\end{tabular}
\end{table}

\subsection{Communication Aware Robot Positioning and Movement Control}
\label{sec:robo_router} In this section, we present a summary of the state-of-
the-art techniques on communication aware positioning and placement control of a
group of robots in order to fulfill data routing and sensing requirements.

\subsubsection{Multi-Robot Sensing} 
One of the main focus of RWSN should be on multiple
robots based sensing with realistic wireless communication constraints.
Note that, every function/sub-problem in an RWSN, such as localization and movement
control, involves some sort of sensing such as RSSI, SINR, or locations of
nodes. However, in this section, we focus on the state-of-the-arts on
multi-robot systems where the main purpose of deployment is to sense an environment.
There exist many works in the field of sensor networks and distributed robotics
that deal with distributed sensing and sensed data collection. However, most of
these works have some idealistic assumptions about either the communication model or
the robot control problem and, thus, not directly applicable in RWSN. 
In this section, we only focus on the existing works on multiple robot based sensing that involves
realistic models for both communication and control.

\textbf{What is already out there?}
In the field of multi-agent sensing, distributed coverage of the area of interest is a very well known topic of research in the contexts of both WSN~\cite{wang2010coverage} and multi-robot systems~\cite{cortes2004coverage}. In the contexts of WSN, coverage control algorithms focus on the placements of static sensor nodes to optimally cover the area of interest. We do not present a survey of this well studied problem. An interested reader is referred to a survey such as \cite{wang2010coverage}. However, most of these works do not use the controlled mobility of the robots to the advantage. On the other hand, there exists a class of coverage control related articles in the field of coordinated robotics that focus on the control and path planning of the robots. Most of these works employ graph theoretic tools such as Voronoi partitions to solve the coverage problem~\cite{schwager2009optimal,choset2001coverage,hazon2008redundancy}. However, these works do not address the problem of collecting and communicating the sensed data effectively. Only recently, a small group of researchers started to look into multi-robot sensing problem from both control and communication point of views. The work of Kantaros and Zavlanos is relevant in this context~\cite{kantaros2014communication}. They looked into the coverage problem of multiple robotic wireless sensors placement by formulating an optimization problem that combines placement optimization with realistic communication constraints and sensing efficiencies of the robotic nodes. In \cite{yan2013communication}, Yan and Mostofi also looked into the problem of robotic path planning and optimal communication strategies in the context of a single robot assisted sensing and data collections. Similar combined path planning and communication optimization in the contexts of multiple robot based system is presented in the works of Ghaffarkhah and Mostofi~\cite{ghaffarkhah2012optimal,ghaffarkhah2012path}. 
The works of Mostofi \et \cite{gonzalez2013cooperative,gonzalez2014integrated,mostofi2013cooperative} on cooperative sensing and structure mapping are also related to this context.
In these works, the authors leveraged multiple pairs of coordinated robots and their RF communication abilities to sense/map unknown structures. To this end, they employed the concepts of compressible sampling/sensing~\cite {candes2006compressive} and the well known propagation properties of RF signals such as path loss and fading~\cite{rappaport1996wireless}.
The work by Le Ny, Ribeiro, and Pappas~\cite{le2012adaptive} also presents an optimization problem that couples motion planning and communication objective for sensing. On similar note, Williams and Sukhatme proposed a multiple robot based plume detection method in \cite{williams2011cooperative}. 
% The work of Jiang and Zefran~\cite{jiang2013coverage} on coverage control is also mentionable.
In~\cite{Gasparri2008}, a new formulation of the multi-robot coverage problem is proposed.
The novelty of this work is the introduction of a sensor network, which cooperates with the team of robots in order to provide coordination. The sensor network, taking advantage of its distributed nature, is responsible for both the construction of the path and for guiding the robots. The coverage of the environment is achieved by guaranteeing the reachability of the sensor nodes by the robots.

\textbf{What would be the potential future research directions?}
To our knowledge, there exist a very few works on robot assisted sensing that involved timely, reliable, and efficient delivery of the sensed data. A major focus of the future should be on such joint optimization of data collection and sensing tasks. Moreover, there might exist many dissimilar robots (each consisting of different sets of sensors) in an RWSN. This requires a simple, unified abstraction in terms of control as well as the sensed data collection. The popular publish-subscribe based frameworks (such as MQTT~\cite{hunkeler2008mqtt}) may be used in such contexts. Moreover, the introduction of controlled mobility has opened up the applications of sparse sensing~\cite{mostofi2013cooperative} which can be exploited for energy efficient, non-redundant sensing.

\begin{table}[!ht]
\centering
\caption{Summary of Multi-Robot Sensing}
\begin{tabular}{ |p{2cm}|p{9cm}| } 
\hline 
% \multicolumn{2}{|c|}{Multi-Robot Sensing} \\ \hline
\multirow{4}{2cm}{Existing Works} 
        & $\bullet$ Multi-agent sensing and coverage algorithms from WSN~\cite{wang2010coverage,cortes2004coverage} \\ 
        & $\bullet$ Distributed coverage control in distributed robotics~\cite{schwager2009optimal,choset2001coverage,hazon2008redundancy} \\
        & $\bullet$ Combined optimization of sensing coverage placement and efficient data routing and collection
        \cite{kantaros2014communication,yan2013communication,ghaffarkhah2012optimal,ghaffarkhah2012path,gonzalez2013cooperative,gonzalez2014integrated,mostofi2013cooperative,le2012adaptive,williams2011cooperative,jiang2013coverage,Gasparri2008} \\ \hline
\multirow{5}{2cm}{Potential Future Directions} 
        & $\bullet$ Sensor data collection abstraction (e.g., publish-subscribe model) for multiple dissimilar robotic systems \\ 
        & $\bullet$ Sparse sensing for energy efficient non-redundant sensing \\
        & $\bullet$ More algorithms on combined optimization of robotic path planning, sensing, and communication quality \\ \hline 
\end{tabular}
\end{table}

\subsubsection{Robotic Router Positioning}

Robotic router/relay placement is a cutting edge topic of research in the field of RWSN.
It mostly concerns the second trend in RWSN research i.e., the use of a robotic network to form a temporary communication backbone or support an existing backbone to improve performance.
The problem of robotic router placements is complex and involves direct relations with many other research pieces of RWSN such as connectivity maintenance, communication link modeling, and localization. 
A robotic router/relay is a device with wireless communication capabilities and controlled mobility. 
Such devices can be employed to form temporary/adaptable communication paths and to ensure robust information flow between a set of nodes that wish to communicate but lack direct links between each other. 
Note that, we use \emph{`robotic router/relay'} to refer to the robotic nodes helping in setting up communication and \emph{`communication endpoints'} to refer to the nodes willing to communicate. 
Robotic relay/router nodes relay messages between such communication endpoints.  
The communication endpoints might have certain communication requirements such as min achievable data rate, high throughput, and lower delay. 
Moreover, the communication endpoints can be mobile or the environment can be dynamic with changing communication link properties. 
The objective of a robotic router placement/positioning algorithm is to place the relays in an optimal manner such that the communication requirements are fulfilled throughout the deployment time, and to adapt with the network dynamics. 
Before moving on, we present a commonly used term in such contexts called \emph{`flow'}. 
\emph{A `flow' is defined as the communication path between a pair of communication endpoints via a set of robotic routers/relays. }
In other words, a set of robotic routers assigned to a flow are dedicated to form and support the communication path between the respective pair of communication endpoints.
Based on the objectives as well as network dynamics, the allocation of a set of robotic routers among different flows may also change over time.
Some of the major components of robotic router placements algorithms are: proper positioning and movement planning of these robotic router, allocation and reallocation of robots among flows as they arrive or disappear, and connectivity maintenance. 
In this section, we present the state-of-the-art techniques on robotic router placements.

%\todo{illustration of the concept via a figure}
\textbf{What is already out there?}
The earlier relevant state-of-the-arts on robotic router placement/movement algorithms are linked to the connectivity maintenance problem. 
A major focus of such methods was to keep a moving target/node connected to a static base station via a set of robots with the assumption of an initially connected network.
%, which is NOT true for a generic robotic router placement problem. Moreover, the proposed methods did not focus on fulfilling communication requirements and followed simple unit disk model.
The work of Stump, Jadbabaie, and Kumar~\cite{stump2008connectivity} is mentionable in this context where either the transmitter node is fixed and the receiver node is moving or vice-versa.  
They developed a framework to control a team of robots to maintain connectivity between a sender and a receiver in such cases. 
Among other state-of-the arts, Tekdas, Yang, and Isler~\cite{tekdas2010robotic} focused on the connectivity of a single user to the base station and proposed two different models based motion planning algorithms. 
One is based on known user motion (user trajectory algorithm) and the other is for unknown-random, adversarial motion of the user (adversarial user trajectory algorithm).  
However, this class of works do not deal with the qualities of the communication links as well as the end-to-end performance.
\rev{De Hoog, Cameron, and Visser~\cite{de2010dynamic, hoog2010selection, hoog2009exploration} have also proposed some techniques for maintaining connectivity to a command center in the context of exploration of unknown environments. In their tree like role based network formulation, the leaf nodes are the `explorers' that explores new frontiers, the root of the tree is the `base-station', and the rest of the nodes are `relays' to keep connectivity between the `explorers' and the `base-station'.}
%They used two metrics in their model: the Fiedler value of the weighted Laplacian matrix and the k-connectivity matrix. 

Over last couple years, a handful of researchers have started exploring the problem with more realistic communication models.
Yan and Mostofi~\cite{yan2010robotic,yan2012robotic} are among these handful of researchers to work on the robotic router problem.
They extended the concept of connectivity maintenance to formulate an optimization problem which considers true reception quality expressed in terms of bit-error rate. 
The goal was to minimize bit-error rates of the receivers for two scenarios of multi-hop and diversity. 
\emph{They also demonstrated that the Fiedler eigenvalue optimization  based approach results in a  performance loss. }
They used an extension of the channel modeling technique introduced in \cite{mostofi2010estimation,malmirchegini2012spatial}. 
In \cite{dixon2009maintaining}, Dixon and Frew presented a gradient based mobility control algorithm for a team of relay robots with the goal of formation and maintenance of an optimal cascaded communication chain between endpoints.
Rather than considering the relative positions of the neighbors, they used SNR of communication links between neighbors.  
They presented an adaptive extremum seeking (ES) algorithm which is employed for operating a distributed controller. 
Goldenberg \et \cite{goldenberg2004towards} presented another distributed, self-adaptive technique for mobility control with the goal of improving communication performance of information flows. 
In their work, they tried to design and analyze a simple system to address three issues: application dependency, distributed nature, and self-organization.
A solution to the problem of computing motion strategies for robotic routers in a simply-connected polygon environment is presented in \cite{tekdas2010maintaining}.
\todo{For a summary of all the relevant references, a reader is referred to Table~\ref{table:router_summary}}.
% Zavlanos, Ribeiro and Pappas~\cite{zavlanos2011framework} have also proposed a
% combined framework for mobility and routing.
% The works presented in \cite{bezzo2011disjunctive,chiu2010anchor} are also mentionable in this context. On a related note, there exist some works in the context of wireless static relay placement~\cite{chattopadhyay2016sequential,ghosh2014you} where a moving agent deploys static wireless relays as it moves from a sink toward a priori unknown source location. These works can be thought of as a special case of robotic relay placements.

%A disjunctive programming approach is presented in \cite{bezzo2011disjunctive}. 
%Chiu and Shen~\cite{chiu2010anchor}, proposed a centralized method of exploring all possible communication links in an environment and then use the collected data to properly position the robotic router. The robots explore all possible communication links between nodes located at different grid points of a known arena. 

As mentioned earlier, the main goals of robotic router placements are to fulfill certain communication requirements such as supporting a set of flows~\cite{williams2013route}, guaranteeing certain performance criterion (say, data rate) to the customers~\cite{gil2015adaptive}, or fixing holes~\cite{tuna2014unmanned}.
In \cite{williams2013route}, Williams, Gasparri, and Krishnamachari presented a hybrid architecture called INSPIRE, with two separate planes called Physical Control Plane (PCP) and Information Control Plane (ICP). 
Their goal was to support a flow based network between multiple pairs of senders and receivers using a group of robots and optimize the overall packet reception rate (PRR) (or the expected number of transmissions, ETX). 
In~\cite{wang2013robotic}, Wang, Gasparri, and Krishnamachari presented a method called robotic message ferrying, where a set of robots literally travel from a source to a sink/destination to deliver data. 
The main objective of this work was the allocation of such robotic router nodes among a set of senders and the optimization of the communication performance such as throughput.
Tuna, Nefzi, and Conte~\cite{tuna2014unmanned} also proposed a centralized method of fixing routing holes (due to absence of nodes or failure of nodes) using a group of robotic routers.
In the proposed method, all nodes communicate to a central server to send the sensed data, which in turn controls the positioning and deployment of a set of UAVs to fix routing holes. 
This algorithm employs geographical routing and Bellman Ford routing algorithms to find the missing nodes/links.
In~\cite{fink2013robust}, Fink, Ribeiro, and Kumar focused on guaranteeing a minimum end-to-end rate in a robotic wireless network. They model the communication channels via a Gaussian process model learned with a set of initially collected data. 
They proposed a stochastic routing variable to calculate an end-to-end rate estimate, which is exploited to find a slack in the achieved rate and the required rate. 
This estimated slack is further used for proper mobility control.
Fida, Iqbal, and Ngo~\cite{fida2014communication} proposed a new metric, called reception probability, and a throughput based route optimization method that employs the new metric. They used Particle Swarm Optimization (PSO)~\cite{kennedy2011particle} for finding the optimal configuration due to non-convexity of the problem. 
Gil \et \cite{gil2015adaptive}, also proposed a method of robotic router placements where the communication demands (in terms of data rates) of a set of clients are fulfilled by another set of robotic routers. The demands are modeled in terms of effective SNR (ESNR) to represent the required rate. Each client is serviced by the router closest to it while the router to router communications are assumed to have a very high capacity.
They used a synthetic aperture radar (SAR)~\cite{curlander1991synthetic} concept based directional signal strength (both amplitude and phase) estimation method and a Mahalanabis distance based cost function for the positioning and path planning of the routers.
Some preliminary works on optimizing SINR of the links, i.e, minimizing the effect of interference is presented in~\cite{GhoshPK16}.
The work of Wang, Krishnamachari, and Ayanian~\cite{wang2015optimism} on robotic router placements in cluttered environments, is also related to this context.
{On a related note, Ghosh and Krishnamachari~\cite{pradipta_tech} showed that there exists a bound on the number of robotic routers we need to deploy to guarantee certain communication requirements in terms of SINR. They also proposed a method of estimating worst case interference and SINR in a flow based robotic router deployment context.}

% Summarizing, the existing state-of-the-arts on robotic router placements focus on different application contexts of RWSN ranging from scenarios with no existing backbone to scenarios with existing backbone but with holes.
% The communication goals are also diverse such as optimization of bit-error rate, guaranteeing certain data rate, and guaranteeing coverage. 

% \todo{Zavlanos ~\cite{zavlanos2011framework} Framework on mobilty and routing.
% }

%\textbf{A Special Case: }
%A special case of robotic router placements lies within the body of work related to robot assisted relay deployment. 
%The goal in this context is to use the robots as the carieers of the relay node. 
%The research challenge in such cases is to optimize the path of the robots for deployment as well as optimize the positioning of the 
% \textcolor{red}{ Perhaps it could be convenient to have in the introduction a discussion of example of integration between mobile robots and sensor networks as discussed in this citation.}

\textbf{What would be the potential future research directions?}
\rev{To our understanding, there exist a lot of research opportunities in this field of research.
First, there is a lack of a physical robotic system based experimentation of the existing works.
One potential direction is to implement some of the promising algorithms and concepts on a real system and perform thorough analysis. To this end, there is a lack of academic open-source robotic network testbeds. Thus building a generic, scalable, adaptable robotic network testbed is another potential direction of research.} Furthermore, most of the solutions are centralized and need to be converted to decentralized methods. Interference among the robotic routers as well as the effect of CSMA or any other channel access are also unaccounted for in most of the existing works. Nonetheless, to our understanding, the main focus of future should be on developing scalable, adaptable physical systems to test out the developed algorithms.

\begin{table}[!ht]
\centering
\caption{Summary of Robotic Router Placements Related Works}
\label{table:router_summary}
\begin{tabular}{ |p{2cm}|p{9cm}| } 
\hline 
% \multicolumn{2}{|c|}{Multi-Robot Sensing} \\ \hline
\multirow{9}{2cm}{Existing Works} 
        & $\bullet$ Tethering a moving object to a base station~\cite{stump2008connectivity,tekdas2010robotic} \\ 
        & $\bullet$ Robot assisted static relay placements \cite{chattopadhyay2016sequential,ghosh2014you} \\
        & $\bullet$ Single communication chain formation with performance guarantee in terms of bit-error rate, number of hops, end-to-end-rate, or SINR~\cite{yan2010robotic,yan2012robotic,dixon2009maintaining,goldenberg2004towards,tekdas2010maintaining,zavlanos2011framework,bezzo2011disjunctive,chiu2010anchor,fink2013robust,fida2014communication} \\
        & $\bullet$ Multiple flow based robotic network~\cite{williams2013route,GhoshPK16} or mesh robotic network placement and performance optimization~\cite{gil2015adaptive,tuna2014unmanned} \\
        & $\bullet$ Use controlled mobility to carry the message from a source to a destination~\cite{wang2013robotic} \\
        & $\bullet$ Integrated framework for network goal oriented mobility control~\cite{zavlanos2011framework,wang2015optimism,williams2013route} \\ \hline
\multirow{5}{2cm}{Potential Future Directions} 
        & $\bullet$  Hardware implementation and evaluation of the existing algorithms \\ 
        & $\bullet$ Include the effects of interference and channel access protocols in the placement optimization problem \\
        & $\bullet$ More decentralized method developments \\ \hline 

% \multicolumn{2}{|c|}{Robotic Router Positioning} \\ \hline
% \multirow{3}{2cm}{Existing Works} 
%         & $\bullet$ Multi-agent sensing and coverage algorithms from WSN~\cite{wang2010coverage,cortes2004coverage,wang2010coverage} \\ 
%         & $\bullet$ Distributed coverage control in distributed robotics~\cite{schwager2009optimal,choset2001coverage,hazon2008redundancy} \\
%         & $\bullet$ Combined optimization of sensing coverage placement and efficient data routing and collection. \hline

\end{tabular}
\end{table}

\subsection{Localization}
%@file j_net_rf_loco.tex
%@brief Network and RF Localization subsection by Jason.
\label{sec:local}
In this section, we present a survey on the state-of-the-arts on localization techniques in the context of RWSNs.
The problem of localization is very well-known in the contexts of sensor networks and distributed robotics. The state-of-the-arts on localization are very mature and require a complete separate survey~\cite{liu2007survey,sun2005signal,guvenc2009survey,amundson2009survey}. In this section, we provide a very brief overview of the existing localization works and point out the works most relevant to the field of RWSN. 
Note that, the concept of \textbf{`localization'} is to locate a node in a deployment arena with respect to a reference frame or a reference location. A commonly used system called the Global Positioning System (GPS) localizes objects in terms of their latitudes and longitudes. 
However, GPS is known to not work properly in cluttered or indoor environments.
Therefore, most of the target application contexts of RWSN require an alternate and efficient localization scheme for indoor environments such as RF based localization.
%Another localization example would be to simply locate a device in room of known/unknown dimensions. 
Moreover, while absolute locations are much important, a relative localization between the nodes in the network is sufficient in many contexts of RWSN. For example, consider a scenario where a group of robotic routers are employed to connect a moving target with a base station. In such contexts, the robots form a chain where each robot positions itself with respect to its neighboring nodes only. Relative positions with respect to the neighboring nodes are enough for a node's movement control decisions in this context.
In this section, we present a summary of the existing literature on the relevant localization techniques.
%Moreover, most of the application contexts of RWSN involved scenarios such as no-line of sights, underground/cluttered environments where traditional GPS or Camera based systems doesn't work, discussed later.

\textbf{Vision and Range Based System:} As mentioned earlier, localization has been a very active field of research in the domain of distributed robotics. The most popular localization systems in the field of robotics employ cameras and range finders. With the help of efficient sampling and filtering algorithms such as Particle Filtering or Kalman Filtering, the camera based systems locate the object in its field of view while range finders provide depth/distance information~\cite{jung2004detecting,schulz2001,papanikolopoulos1993visual,lindstrom2001detecting} .
In order to deal with the movements as well as errors, some researchers use temporal snapshot of the targets~\cite{markovic2014moving,prassler2000tracking}.
{However, any camera/vision based approach has many limitations such as the
visibility requirement, limited field of vision of traditional cameras, larger
form factor of the robots, and costly image processing software requirements.
On the other hand, while the laser range based methods do not suffer from the
visibility problem, they are limited to direct line of sight between the target
and the tracker and require complicated processing. }

\textbf{RF Based System:}
As an alternative to GPS and vision based localization systems, wireless sensor network researchers have proposed a variety of Radio Frequency (RF) based localization systems. 
As mentioned earlier, there exist a range of survey papers with the sole focus on such RF localization techniques \cite{liu2007survey,sun2005signal}.
Briefly speaking, the existing localization techniques employ either of the following aspects: Direction of Arrival (DoA), Time of 
Arrival (ToA), Time Difference of Arrival (TDoA), Received Signal Strength (RSS), and proximity.
The typical underlying technologies used to realize these techniques are RFID, 
WLAN, Bluetooth, and ZigBee. 
Liu \et \cite{liu2007survey} provided a great outline and classifications of general wireless localization techniques and systems. They outline most of the performance metrics used in traditional RF based localization as follows: accuracy (mean error), precision (variance, or distribution of accuracy), complexity, robustness, scalability, and cost. 
With the recent trends of networked robots, it is of interest to the research community to look  at these performance metrics from the perspectives of RWSN. 
The difference between an RWSN and a WSN with static and mobile actuating sensor nodes is that robots are more dynamic in position and require greater performance and flexibility in localization due to the larger range of tasks the robots are expected to perform.  
%In the following, we discuss the existing state-of-the-arts on RF based localization in robotics.

\textbf{RFID Based System:}
Over last two decades, many researchers have peered into the use of
RFID tags because of their \emph{low cost and power (or zero power for passive tags)}.
A confined deployment arena for a team of robots can be fitted with a mass deployment of RFID tags. Then, we can localize any RFID/RF device carrying robots in that arena~\cite{yamano2004self,hahnel2004mapping}. Zhou \et present a survey of existing research and deployments of RFID tags for
localization in \cite{zhou2009rfid} which shows its usefulness in robotics. 
However, one large quirk of RFID tags is the static nature of the tag placements and limited tag functions and information. The benefits of RFID tags manifest when meticulous planning or post-deployment positioning (using a robot) is done.
\textcolor{black}{In~\cite{Brass:2011}, a multi-robot exploration of an unknown graph  describing a set of rooms connected by opaque passages is considered.  In particular, the authors demonstrate how in this framework, which is appropriate for scenarios like indoor navigation or cave exploration, communication can be realized by bookkeeping devices, such as RFID tags, being dropped by the robots at explored vertices, the states of which are read and changed by further visiting robots.} 
 As an alternative, RFID tags can be replaced or complemented by WSNs, which have greater capabilities and flexibility.

\textbf{Wireless AP Based System:} With the ubiquity of WLAN access points and wide availability of wireless sensor nodes, the research community has investigated the use of a network infrastructure to position and navigate robots. The work of Ladd \et \cite{ladd2005robotics} illustrated the
feasibility of using commercial off-the-shelf radios and radio signal strength measurements as a robust locater of robots. To enhance the overall performance of RSS based localization, researchers have investigated RF scene analysis or fingerprinting as a viable option for indoors. Ocana \et \cite{ocana2005indoor} presented such a robot localization system that starts with a semi-autonomous method to fingerprint indoor environments using a robot. Many other research teams followed up with works concentrated on RF surveying with directional antennas on  a robot, such as in \cite{song2009monte,venkateswaran2013rf,palaniappan2011autonomous,dantu2009relative,deyle2009rf}. We encourage readers to also refer to our section on RF mapping (Section~\ref{sec:rf_mapp}) to complement the RF scene analysis.

\textbf{Distributed and Cooperative System:}
Distributed and cooperative network and RF based localization in a dynamically moving network of robots is still not a fully understood area. Wymeersch~\cite{wymeersch2013communications} showed the positive impacts of cooperative localization on achieving more  complex tasks with a team of robots. 
This further motivates the need for a better understanding of cooperative network and RF based localization to see if we can enhance the functionality of RWSNs. 
In \cite{koutsonikolas2006cocoa}, Koutsonikolas \et delved into the problem of cooperative localization, but the authors assume there are a subset of robots that carry extra sensors, including GPS, to aid in the overall system. The rest of the robots carry 802.11 radios, only using beacons to determine proximity. 
On that node, the work of Zickler and Veloso~\cite{zickler2010rss}focus on relative localization and tethering between two robots based on the received signal strengths. They opt for a discrete grid based Bayesian probabilistic approach. In their system, a locator node moves to different relative positions with respect to the node being localized to collect multiple RSS values while they communicate their odometer readings. 
\textcolor{black}{In~\cite{Filoramo:2010}, Filoramo \et describe an RSSI-based technique for inter-distance computation in multi-robot systems. In particular,  for a team of robots equipped with Zigbee radio transceivers, they  propose a data acquisition technique which relies on spatial and frequency averaging to reduce the effect of multi-path for both indoor and outdoor environments. Furthermore, they show how the proposed data acquisition technique can be used to improve a Kalman Filter-based localization approach.} Another RSSI based relative localization system is proposed by Oliveira \et \cite{oliveira2014rssi} which also relies on pairwise RSSI measurements between the robots. To improve the performance and accuracy of the localization, they apply Kalman filter and the Floyd–Warshall algorithm. One of the most recent significant work on relative localization is presented in~\cite{vasisht2016decimeter} which applies MIMO based system for a single node based localization. These methods are particularly relevant to the field of RWSN. 

\rev{
%Among the modern works in RWSN, the work of  
%Yan and Mostofi~\cite{yan2013co} is worth mentioning. They
%explore the co-optimization of communications and motion planning of robotic operations. 
%Hsieh \et \cite{hsieh2008maintaining} studied the creation and use of RSS maps to maintain communication links.
%These works highlight the new trend of location-aware communications in RWSN. Leveraging proper localization techniques for robotic applications, location-awareness can revolutionize the way we look at communications research for robotics. 

\textbf{What would be the potential future research directions?}
Most of the current robotics network related research are performed in artificial fixed indoor environments as there exist costly camera based solutions (such as VICON system~\cite{vicon}) to provide millimeter level accuracy required for the experimentation. We believe that RF localization will be able to help extend such experiments to truly indoor cluttered environments with much lower cost than deploying camera systems. However, most of the existing RF solutions are also not portable e.g., they require either a fixed infrastructure with RF beacons, a map of the environment, etc. Therefore, the future direction of localization relation research should be focused on developing \textbf{portable scalable} at-least centimeter level accurate RF localization systems that can be quickly deployed on demand and can be remove easily. Moreover, the frameworks should be portable to deploy in real-world applications such as in firefighting. In that context, another potential direction is towards `self sufficient' RWSN i.e., the network will not require help from any existing infrastructure to perform the assigned tasks. Further, researchers should study different localization properties such as accuracy, complexity, and communication requirements in the context of RWSN where a robot's performance directly relies on localization e.g., in the context of tracking and following a firefighter while providing connectivity to a command center.

}
\begin{table}[!ht]
\centering
\caption{Summary of Localization Related Works}
\begin{tabular}{ |p{3cm}|p{2cm}|p{6cm}| } 
    \hline
    Localization Algorithm Type/Class & References & Comments \\ \hline
    \multirow{6}{3cm}{Vision and Range Based System} & \multirow{6}{2cm}{\cite{jung2004detecting,schulz2001,papanikolopoulos1993visual,lindstrom2001detecting,markovic2014moving, prassler2000tracking} }
                         & $\bullet$ Popular in robotics\\ 
                   &     & $\bullet$ High accuracy\\ 
                   &     & $\bullet$ Many off-the-shelf solutions are available\\ 
                   &     & $\bullet$ Requires heavy computation\\ 
                   &     & $\bullet$ Requires line of sight \\ 
                   &     & $\bullet$ Large form factor\\ \hline 

    \multirow{8}{3cm}{RF Based System} & \multirow{8}{2cm}{\cite{liu2007survey,sun2005signal} }
                         & $\bullet$ Popular in WSN community\\
                   &     & $\bullet$ Uses RF signal properties (like signal strength or time of arrival) \\ 
                   &     & $\bullet$ Lower accuracy compared to camera based system\\ 
                   &     & $\bullet$ Low cost and low computation\\ 
                   &     & $\bullet$ No line of sight requirement \\ 
                   &     & $\bullet$ Small form factor\\ \hline            
    \multirow{6}{3cm}{RFID Based System} & \multirow{6}{2cm}{\cite{yamano2004self,hahnel2004mapping,zhou2009rfid,Brass:2011} }
                         & $\bullet$ Low cost and power (or zero power for passive tags)\\
                   &     & $\bullet$ Requires static placements of the tags in the deployment arena \\ 
                   &     & $\bullet$ Requires pre-deployment of the localization infrastructure\\  \hline            
    \multirow{5}{3cm}{Wireless AP Based System} & \multirow{5}{2cm}{\cite{ladd2005robotics,ocana2005indoor,song2009monte,venkateswaran2013rf,palaniappan2011autonomous,dantu2009relative,deyle2009rf}}        & $\bullet$ Use pre-deployed or existing WLAN access points\\
                   &     & $\bullet$ Use commercial off-the-shelf radios \\ 
                   &     & $\bullet$ Requires pre-mapping or fingerprinting of the deployment region\\  \hline         

    \multirow{8}{3cm}{Distributed and Cooperative System} & \multirow{8}{2cm}{\cite{wymeersch2013communications,koutsonikolas2006cocoa,zickler2010rss,Filoramo:2010,oliveira2014rssi,vasisht2016decimeter}} 
                         & $\bullet$ A subset of anchor robots have GPS or other localization capabilities\\
                   &     & $\bullet$ Rest of the robots localize themselves relative to the anchor nodes  \\ 
                   &     & $\bullet$ Sometimes odometer readings are combined for better accuracy   \cite{zickler2010rss}\\ 
                   &     & $\bullet$ Some types of filtering can be involved to improve performance\\ \hline  \hline 

     \multirow{3}{3cm}{Future Directions} &  \multicolumn{2}{l|}{$\bullet$ Co-optimizations of localization accuracy and communication goal} \\
                   &  \multicolumn{2}{l|}{ $\bullet$ Develop cheap, scalable, high accuracy system}  \\ 
                   &  \multicolumn{2}{l|}{$\bullet$ Study the trade-offs among accuracy, complexity, and cost}\\ 
                        \hline

\end{tabular}
\end{table}
\section{RWSN Network Stack Layer Analysis:}
In the last section, we categorized the exiting state-of-the-arts in the field of RWSN according to the key research problems in focus. 
The majority of the works in RWSN are focused on \emph{RSS modeling and mapping}, \emph{connectivity maintenance}, \emph{routing algorithms}, \emph{communication aware robot placements}, \emph{connectivity maintenance}, and \emph{localization}. 
However, one key thing missing in the last section is how all these works fit in the contexts of traditional networking concepts.
Traditionally, the protocols relevant to a network (say sensor network) are developed by following the well-known layered network stack architectures such as OSI model or Internet model.
Ideally, we would want the same for RWSN.
However, we discover that the layered structure in an RWSN device does not follow the traditional norm. 
Rather, it mostly relies on inter-dependencies between layers. 
To understand the layering requirements, we first analyze the existing works explained in the last section according to the five layered Internet protocol stack: Physical layer, MAC Layer, Network layer, and Transport layer, Application layer. Next, we discuss about some potential unified architectures for RWSN.

\subsection{Internet Model for Network}
\label{sec:int_rwsn}

\subsubsection{Physical Layer}
\label{sec:phy}
The physical layer in a traditional network deals with the physical communication between nodes. The function of physical layer includes but not limited to representation of bits, controlling data rate, synchronization between transmitter and receiver, defining the communication interface, and controlling the mode of communication such as simplex or duplex. Therefore, research on topics such as communication hardware, physical medium and communication technologies (e.g., RF and Bluetooth), and modulation-demodulation of signal falls under this category.
%In case RWSN, the works related to RSS Measurement, RF Mapping and Position Control to improve link quality are related to the signal strengths of the physical signal fall under this category.
There exist many sorts of wireless communication technologies such as RF, Bluetooth, and Sonar for communication among robots.
For obvious reasons, the dominant technologies for above-ground communication are RF communication techniques~\cite{gage1998network,fung1994position, wilke2001flexible,thompson2015robot}. 
For short range communication, some networks of robots use bluetooth \cite{fai2002bluetooth} and infrared.
The above-ground radio frequency based methods are not applicable for acoustic communication (e.g., underwater communication) due to reasons like high loss exponents and fading due to turbulence. 
The most common acoustic communication techniques are special RF communication \cite{heidemann2006research} and sonar. 
While RF based communication still remains the mainstream for RWSN system, the introduction of robots and controlled mobility have also opened up  some new but relatively unexplored communication methods. On that node, Ghosh \et \cite{ghosh_madcom} have presented a proof-of-concept of a new method of communication that employs the location and the motion pattern of a communicating robot as the communication signal. 

Most of the works on RSS measurements, RF mapping, and position control to improve link qualities fall under the purview of the physical layer since these works are directly linked to the signal strengths variations over the area of interest. 
%However, as briefly mentioned earlier, most of the application contexts of RWSN relies on a rather cross-layer approach. For example, 
%classical solutions to the robotics router placement problem incorporate physical signal strength variations over the space and the received signal strengths at each node along with information about routing paths.. 
%These methods properly combine controlling aspect of robots with the signal strength knowledge to create and maintain a efficient communication path for a information flow.
%So these methods can be considered as a cross-layer methods  that falls under both physical layer and routing layers.
On a similar note, most of the RF based localization techniques use properties (such as signal strength) of the communication signals (RF or ultrasound) and, thus, also fall under the purview of the physical layer.
Among the emerging research domains related to physical layer of an RWSN, the concept of distributed  MIMO implementation using robots is promising.
In \cite{zhang2006twinsnet},  Zhang \et presented a cooperative MIMO like communication structure in mobile sensor/robotic networks. They subdivide a network into twin sub-networks where each transmitter node pairs itself with another node for transmitting cooperatively in a MIMO like fashion.  

{
Another class of work that also partly falls under the purview of physical layer lies within the communication aware robot positioning and movement control related works where the robots adapt their positions to optimize the quality of the communication links. In such cases, the robots employ the physical layer information such as RSSI~\cite{GhoshPK16} or data rate~\cite{gil2015adaptive} to control the movements of the robots acting as relays/routers/sensors.}
% In such contexts, we need to include another layer called movement control layer which works in close integration with the physical layer.} 

In summary, we can state that a major focus of most of the existing state-of-the-arts on RWSN has been towards the physical layer.
%Also in their technique other nodes act as relays and place themselves accordingly. 

%
%Despite of this huge amount of works, a lots of rooms are still left to be done along with proper practical implementation and analysis of all this methods.
%In the area RSS modeling and mapping, more realistic models are needed which should include the effects of interference on the communication along with proper path loss, fading and shadowing effects.
%An ideal wireless networks among wireless robots should be able to update its models and measurement over the motions.
%For robotic router related problems, most of the works do not include the inter-flow communications and interferences.
%Also distributed MIMO communication using robots is at the very early stage of research without much documentation. 
%So, this area also has a huge potential as a research topic.
%

\subsubsection{Media Access Control Layer}
Briefly speaking, Media Access Control (MAC) protocols deal with proper distributed access of the physical medium among nodes, node to node communication, framing, and error corrections. 
While most of the classical MAC layer modules and protocols~\cite{rappaport1996wireless} are applicable to RWSN, to our knowledge, there exist only a few MAC layer protocols designed specifically for networking between robots. 
Related to media access protocols, CSMA/CA is an obvious choice for RWSN due to its ubiquitous properties such as randomness and scalability.
On the other hands, TDMA and FDMA systems can also be modified for RWSN with the extra feature of controlled mobility. 
On that note, Hollinger \et \cite{hollinger2011communication} presented a MAC protocol for robotic sensor networks in acoustic environments.
They proposed a three-phased method based on TDMA with acknowledgments. The phases are Initiation, Scheduling, and Data transfer, respectively. 
There also exist works related to mobile WSN which use predicted mobility patterns, such as pedestrian mobility or vehicular mobility, of both source and sink to design efficient application context specific MAC protocols. 
Some examples of such MAC protocols are: MS-MAC~\cite{pham2004adaptive}, M-MAC~\cite{ali2005mmac}, M-TDMA~\cite{jhumka2007design}, MA-MAC~\cite{zhiyong2010mobility}, MobiSense~\cite{gonga2011mobisense}, and MCMAC~\cite{nabi2010mcmac}.
These works show that many difficulties arises as a result of mobility (mainly uncontrolled mobility) such as random variations in link quality and frequent route changes. 
To deal with these problems of mobility, researchers proposed a class of methods like negotiation based rate-adaptation and handover by transmitter. 
For more detailed survey on such techniques, the reader is referred to \cite{dong2013survey}. 
In contrast, the mobility of the nodes in RWSN are controlled and, thus, can be exactly known or predicted with higher accuracy. 
This opens up a new domain of research where the mobility controller and the MAC protocol could work in an integral manner to optimize the utilization of the radio resources. 
  
In summary, while the existing MAC protocols are applicable to RWSN, there is a lack of MAC layer protocols specifically designed and optimized for RWSN. 
As mentioned earlier, one of the future directions would be to incorporate  mobility control with MAC access protocols for better performance. 
Another future direction would be to use the knowledge of the MAC protocol to estimate signal properties such as Interference and SINR~\cite{pradipta_tech}, and to control the link properties such as interference~\cite{ning2014minimizing}.
  
%  \todo{ need to check if there has been any recent works}

 \subsubsection{Network Layer}
 
Network Layer in a traditional wireless network deals with the packetization of data as well as routing of the packets from source nodes to respective destination nodes.
One of the main application contexts of RWSN is to maintain a temporary communication backbone to support data flow between communication endpoints. 
Thus, a major focus of RWSN related works till date has been on developing network layer protocols and combining controlled mobility with routing of packets. 
All the routing related works presented in Section~\ref{sec:routing} directly fall under this category for obvious reasons. 
Similarly, all the works on communication aware robot placements (discussed in section~\ref{sec:robo_router}) also fall under the purview of network layer.
However, as mentioned earlier, the concept of layering in RWSN is slightly vague as it relies on cross-layer dependencies such as dependencies between the physical layer and the network layer. 
This is mostly apparent in the contexts of robotic router placement algorithms that deal with the placements of the robotic nodes (based on the physical layer information) to optimize the end-to-end path from a source to destination as well as to optimize each link~\cite{sliwa2016bat,williams2013route}. 
%  There is lots of area for improvements. In most of the work the robots are assigned to a particular flow instead and each robot help only one flow. 
%  There lacks sufficient works that deals with using some nodes as common between flows.
All connectivity related works presented in Section~\ref{sec:connectivity} also fall under this category as the key goal in such connectivity maintenance algorithms is to guarantee the existence of a communication path between every pairs of nodes in the network. 
Without connectivity, the network might be segregated into smaller sub-networks and won't be able to fulfill the routing goals.
   
\subsubsection{Transport Layer}
   
%   \todo{ do more thorough research. saved for the end}
%   There is no significant work on transport layer related areas in RWSN. Most of the robotics researchers use UDP as the transport layer protocol. Original TCP is not suitable for RWSN due to many reasons such as longer delay to provide reliability. Similarly UDP is also not suitable for RWSN due to mobility of the robots. Most of the available protocols can not guarantee the timely delivery of the control messages thereby hampering the operation of the network. None them also use the controllable mobility into account. There is a need of novel protocol that is suitable and efficient for RWSN.   
In the field of RWSN, the researchers are yet to significantly focus on the transport layer protocols. 
Till date, researchers employed traditional transport layer protocols such as TCP, UDP, or some MANET transport layer protocols for robotic networks. 
However, unlike MANET, robotic networks have an extra feature of controllable mobility which provides an extra dimension. 
But, there is no significant work on controlling mobility for improved performance of transport layer protocols.
Conversely, controllability requires highly reliable, {low delay}, and error free communication between robots, which is not possible using original TCP or UDP and requires some special transport layer protocol. 
In this section, we present a brief overview of the existing transport layer protocols for RWSN.
% Also there are some protocols for remotely operating robots. But most of these works are focused on congestion control. 

Among the state-of-the-arts, the work of Douglas W. Gage on the MSSMP Transport layer protocol (MTP)~\cite{gage1998network} is mentionable. This protocol is based on the Reliable Data Protocol (RDP). 
The RDP is much more effective and appropriate service model for mobile robot applications than TCP. 
There are many features of RDP which are more useful in RWSN such as more communications bandwidth than TCP and simpler in terms of implementation. 
But the interface to the application layer in MTP is similar to TCP, i.e., based on socket like API.
There is also couple of congestion control protocol designed for tele-operation of robots such as  Trinomial method\cite{liu2005end},  Real-Time Network Protocol (RTNP) \cite{uchimura2004bilateral}, and Interactive Real-Time Protocol (IRTP) \cite{ping2005transport}. 
All these methods are not directly related to the RWSN but can be ported. Similarly, there exists a group of works on Transport layer for MANET  \cite{holland2002analysis,harras2006transport}. 
For detailed overview on existing congestion control protocols of mobile ad-hoc network, an interested reader is referred to~\cite{lochert2007survey}.
In summary, the transport layer related research on RWSN requires significant attention in future with a major focus on reliability and delay performances.

\subsubsection{Application Layer}

Many of the existing works related to RWSN are actually related to the application layer.
{
In some of the target application contexts, the robots in an RWSN need to make informed movement and communication decisions in order to work in a cooperative manner. 
For example, in order to form a communication relay path between a pair of communication endpoints, the relay robots need to process a combination of information such as neighboring node locations, flow endpoint requirements (say, a minimum data rate), current link status, and expected interference power to adaptably and optimally position the relays. According to the layered hierarchy, all these processing and decision making should be done in the application layer to keep the system modular. 
%Note that, these movement control decisions are different than the function movement control layer mentioned in the context of physical layer.  A simple example would be, say the movement control decision is to move to a location coordinate while the function of the movement control layer would be translate that into a real motion via translation and rotation of the robot.
For example, in the route swarm work~\cite{williams2013route}, the Information Control Plane (ICP) takes care of making decision regarding state changes of the robots as well as the allocation of robots among different flows. Thus, the ICP in that architecture is mostly implemented in the application layer. 
In the same manner, all the robotic router related works are partly/fully dependent upon the application layer. 
}

%We won't delve into application layer related works on RWSN. The concept of overlay networks can be easily applied to this field.
Another mentionable field related to the application layer is cloud robotics. Cloud robotics basically uses an application layer abstraction of a heterogeneous network of robots to perform a group of tasks. Cloud robotics provides a unified scalable control platform for a group of heterogeneous robots.  The work of Du \et \cite{du2011design} is among the most promising works in this field of research. Quintas \et \cite{quintas2011cloud} also proposed a related architecture. In \cite{kamei2012cloud}, Kamei \et presented a detailed study of the advantages, concerns, and feasibility in Cloud Networked Robotics. In this paper, we do not delve into cloud robotics as it does not directly fall under the RWSN research domain.
\todo{
\begin{table}[H]
\centering
\caption{Summary of Relevant Keywords for the RWSN Layering Architecture}
\label{table:network_Stack}
\begin{tabular}{|p{2cm}|p{2cm}|p{7cm}|}
\hline
Layer & References & Related Keywords\\ \hline
Physical Layer & \cite {gage1998network,fung1994position,wilke2001flexible,thompson2015robot,fai2002bluetooth,heidemann2006research,ghosh_madcom,zhang2006twinsnet} & RSS Measurement, RF Mapping, Localization, Distributed MIMO, Connectivity, Robotic Router Placement, Communication Aware Robot Positioning\\ \hline
MAC Layer & \cite{ rappaport1996wireless,hollinger2011communication,pham2004adaptive,ali2005mmac,jhumka2007design,zhiyong2010mobility,gonga2011mobisense,nabi2010mcmac,dong2013survey,pradipta_tech,ning2014minimizing} & Scheduling, CSMA, TDMA, FDMA\\ \hline
Network Layer & & Routing, Robotic Router Placement, Communication Aware Robot Positioning, Connectivity \\ \hline
Transport Layer & \cite{gage1998network,liu2005end,uchimura2004bilateral,ping2005transport,holland2002analysis,harras2006transport,lochert2007survey} & Delay, Real-time communication, UDP, TCP \\ \hline
Application Later & \cite{williams2013route, du2011design, quintas2011cloud,kamei2012cloud} & Connectivity, Positioning, Robotic Router Placement, Communication Aware Robot Positioning, Cloud Robotics \\ \hline
\end{tabular}
\end{table}
}

In summary, there is no clean way of classifying the existing works into the layered architecture. Rather, each of the problems and solutions belong to multiple layers. Moreover, we need a new layer/module to deal with the mobility control.
All these lead us to believe that maybe we need a new architecture for RWSN that builds upon the existing layered network stacks, discussed in the next section.
Note that we present a summary of our network protocol stack related discussion in Table~\ref{table:network_Stack}.

\subsection{An Unified System Architecture For RWSN}
Based on our analysis in Section~\ref{sec:int_rwsn}, we find that the existing works in RWSN do not fit well in the Internet model of networking stack.
Rather, most of the solutions in RWSN require inter-layer dependencies.
For example, a robotic router placement algorithm relies on the physical layer estimation models which in turn rely on some knowledge about the relay node positions and the network graphs.  
Moreover, we need to have a control layer to combine the network goals with the movement of the robots. 
Thus, we require a new system hierarchy for RWSN where the existing network architecture can be kept intact to the most extent.
On that note,  Williams, Gasparri, and Krishnamachari~\cite{williams2013route} have proposed an architecture with two planes: the Information Control Plane (ICP) and the Physical Control Place (PCP) where the ICP takes care of the networking as well as high level movement decisions and the PCP takes care of the movements. 
There also exist software architecture solutions for a system of multiple robots such as ALLIANCE\cite{parker1994alliance} and CLARAty \cite{volpe2001claraty}.
The works of Arkin and Balch\cite{arkin1998cooperative} and Stoeter \et \cite{stoeter2000robot} are also relevant in this context.

\begin{figure}
\centering
\includegraphics[width=0.8\linewidth]{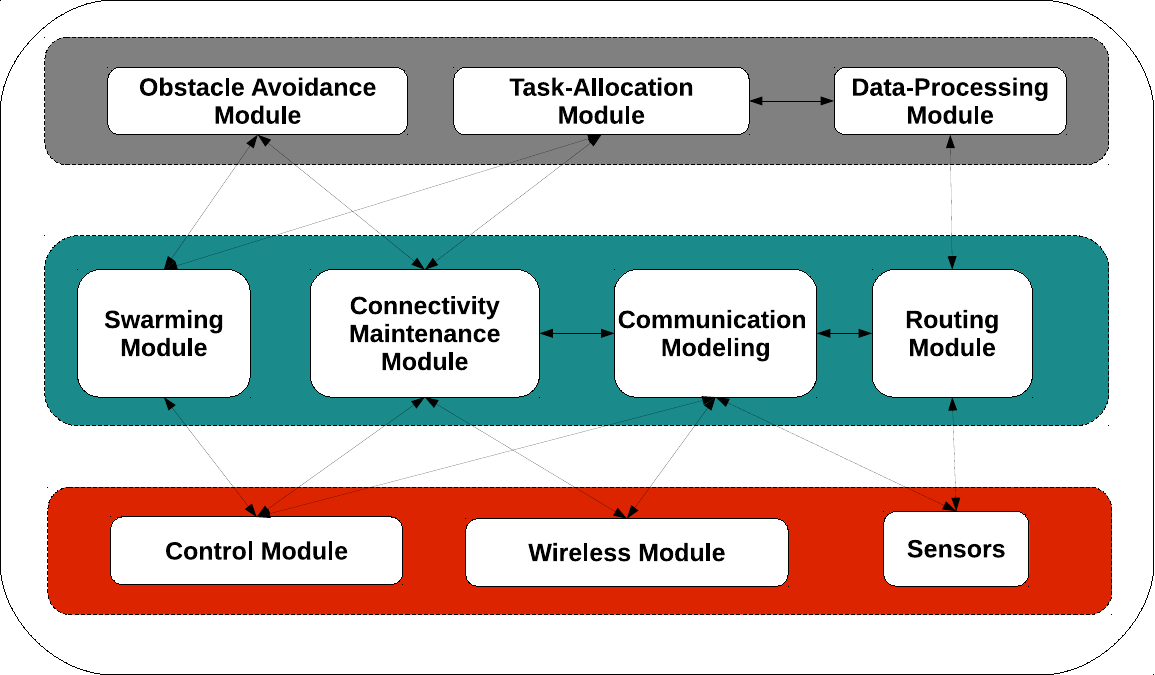}
\caption{Illustration of a Unified Architecture}
\label{fig:unified_archi}
\end{figure}

However, the concept of a unified system architecture is one of the relatively unexplored domain of research in RWSN.  
Most of the existing literature emphasize only certain aspects of system challenges instead of focusing on the networked robot system as a whole.  Thus, there is a need of a \textbf{base, decentralized, realistic system framework using realistic communication models} that can autonomously control an individual robot as well as a group of robots in any kind of RWSN application contexts. \textbf{The term \emph{framework} here refers to a collective set of system modules such as movement control, sensors and connectivity maintenance with necessary interconnections, as illustrated in Figure~\ref{fig:unified_archi}.} This base framework should have \textbf{plug \& play} flexibility as well, i.e., any extra module pertinent to a specific requirement can be added or removed. 
In Figure~\ref{fig:unified_archi}, we present a sample, illustrative architecture for RWSN, based on our understanding.
%Note that, this is just an illustration of what an unified architecture would look like.

\section{Collaborative Works on Networked Robots}

There are many projects on collaborative robotics with different goals in focus. 
Among them, Ubiquitous Robotics network system for Urban Settings (URUS) project (http://urus.upc.es) \cite{sanfeliu2006ubiquitous}, Japan's NRS project \cite{sanfeliu2008network}, Physically Embedded Intelligent Systems Ecology \cite{saffiotti2008peis} project, DARPA LANdroids program \cite{mcclure2009darpa}, Mobile Autonomous Robot Systems (MARS) \cite{chaimowicz2005deploying}, Mobile Detection Assessment and Response System (MDARS) \cite{gage1998network} are the important ones. Among other projects, the swarm-bot project by École Polytechnique Fédérale de Lausanne (EPFL) (\cite{dorigo2005swarm}, \cite{mondada2005superlinear}), the NECTAR Project~\cite{nectar} by Filippo Arrichiello and Andrea Gasparri, and I-Swarm project (\cite{seyfried2005swarm}, \cite{woern2006swarm}) by Karlsruhe are the significant ones.  

Since we are mainly interested in network related research in RWSN, the main project that falls in our category is the the DARPA LANdroids program. 
This is one of the recent projects undertaken on RWSN. Tactical communication enhancement in urban environments is the main goal of this program \cite{mcclure2009darpa}. Towards this goal, the researchers tried to develop pocket-sized intelligent autonomous robotic radio relay nodes that are inexpensive. One of the serious communications problems in urban settings is multipath effect. LANdroids are envisioned to mitigate the problem by acting as relay nodes, using autonomous movements and intelligent control algorithms. LANdroids will also be used to maintain network connectivity between dismounted war-fighters and higher command by taking advantage of their co-operative movements.

On the other hand, there are some industrial projects that also fall under the purview of RWSN such as Facebook's Aquila project~\cite{acquila} and Google's project Loon~\cite{prjctloon}. There are also some open source projects on swarm robotics such as Swarmrobot~\cite{swarmrobot} and Swarm-bots~\cite{swarmbot}. 
Swarming Micro Air Vehicle Network (SMAVNET) is a related project by EPFL where a swarm of UAVs are envisioned to be used to create temporary communication networks.

\section{Summary and Conclusion}
\rev{ The main aim of this chapter was to identify and define a new field of research, RWSN, and provide a starting point to the new researchers. Briefly speaking, a RWSN consists of a group of controllable robots with wireless capabilities that is deployed with the goal of improving/providing a portable wireless network infrastructure in application with need of sudden and temporary wireless connectivity such as in a search and rescue mission or in a carnival.
While there exist a range of relevant state-of-the-arts, the application of controlled mobility to the advantage of wireless communication is still an open area of research. 
%There is also a lot of opportunities for system researchers to develop and standardize a modular architecture for RWSN as well as to develop relevant, cheap, and scalable hardware.
However, like every new field of research there are some challenges in RWSN research. Some of the known challenges are: (1) lack of programmable, scalable RWSN testbeds for implementing and validating concepts, (2) lack of  good venues to publish research  (there is only a handful of new workshops and conferences that focus on RWSN), (3) because of the inter-disciplinary nature of this field it requires the researchers to have knowledge on a vast range of topics such as robotics, control, communication, embedded systems, and networks. 
Nonetheless, based on all the discussions in this chapter, it is evident that RWSN is an emerging and promising piece of technology with limitless possibilities. 
Some of the known promising ongoing and future directions of RWSN related research can be listed as follows.

\begin{longtable}{| p{.20\textwidth} | p{.80\textwidth} |} 
	\hline
	Systems  					&   $\Box $ Build a full fledged low-power reusable RWSN testbed \\
								&   $\Box $ Implement and analyze promising theoretical concepts on a real system \\ 
								& 	$\Box $ Extensive measurements in real environments (such as mines), identify the RF properties, and formulate communication models and emulators \\ \hline

	Modeling and Mapping 	&  $\Box $ Build a mathematical or systemic model for Interference and SINR estimation in a RWSN \\
								&  $\Box $ Incorporate the effects of MAC protocols such as CSMA into interference estimations \\
								&  $\Box $ RF based online mapping of an unknown environment \\	\hline
								
	Routing 					&  $\Box $ Develop routing algorithms with guaranteed lower delay but higher reliability \\
								&  $\Box $ Apply existing MANET protocols in the context of RWSN \\
								&  $\Box $ Include controllability of the nodes in the routing decision where a bad link can be potentially improved via small movements \\ \hline

	Connectivity maintenance & $\Box $ Realistic communication model based connectivity control  \\ \hline
	Robotic Router			&	$\Box $ Optimization of router placements  with realistic SINR models     \\ 
								&  $\Box $ Guaranteed communication performance such as min achievable data rate by placing robotic routers between TX-RX pairs \\ 
								& $\Box $ Robot based communication link repair \\ 
								& $\Box $ Robotic message ferrying related research with more focus on the tradeoff between movement energy consumption cost and the payoff from good performance or timely message delivery \\  \hline
								
	Localization 				&  $\Box $ Build a portable RF based localization system with at least centimeter level accuracy \\
								&  $\Box $ Focus more on relative localization than absolute localization		\\ \hline
	Network Stack			&  $\Box $ Multiple robot based co-operative MIMO 			\\
								&  $\Box $ MAC protocols with mobility control, engineered specifically for RWSN\\
								&  $\Box $ Transport layer protocols (alternate to UDP or TCP) engineered for RWSN \\ 
								& $\Box $ Unified system architecture for RWSN \\ \hline
\caption{Ongoing and Future Research Directions} % needs to go inside long-table environment
\label{tab:future_res}
\end{longtable}

}

%We to include all significant works on the respective fields. But there might be some works left to be included. Our future goal is to develop new wireless network standards for networked robots to solve the problems better that current available methods.  

\bibliographystyle{unsrt}
\bibliography{WNR}
    
\end{document}